%% file: M2P_v1.tex
\documentclass[lettersize,journal]{IEEEtran}
\usepackage[hidelinks]{hyperref}
\usepackage{amsmath,amsfonts}
\usepackage{algorithmic}
\usepackage{algorithm}
\usepackage{array}
\usepackage[caption=false,font=normalsize,labelfont=sf,textfont=sf]{subfig}
\usepackage{textcomp}
\usepackage{stfloats}
\usepackage{url}
\usepackage{verbatim}
\usepackage{graphicx}
\usepackage{cite}
\usepackage{colortbl} 
\usepackage{xcolor}
\usepackage{amssymb}
\usepackage{booktabs}      
\usepackage{multirow}      
\usepackage{graphicx}      
\usepackage{makecell}
\usepackage{paralist}
\usepackage{pifont}
\usepackage{caption}

\newcommand{\real}{\mathbb{R}}
\newcommand{\bF}{\mathbf{F}}
\newcommand{\bff}{\mathbf{f}}
\newcommand{\bp}{\mathbf{p}}
\newcommand{\bQ}{\mathbf{Q}}

\newcommand{\cmark}{\ding{51}}%
\newcommand{\xmark}{}%

\newcommand{\CUT}[1]{}

\newcommand{\yty}{} 

\begin{document}

\title{M2P: Improving Visual Foundation Models with Mask-to-Point Weakly-Supervised Learning for Dense Point Tracking}

\author{Qiangqiang~Wu,
        Tianyu~Yang,
        Bo~Fang,
        Jia~Wan,
        Matias~Di~Martino,
        Guillermo~Sapiro,~\IEEEmembership{Fellow,~IEEE,}
        and~Antoni~B.~Chan,~\IEEEmembership{Senior Member,~IEEE}
\thanks{Q. Wu, B. Fang, and Antoni B. Chan are with the Department of Computer Science, City University of Hong Kong, China (e-mail: qiangqwu2-c@my.cityu.edu.hk; bofang6-c@my.cityu.edu.hk; abchan@cityu.edu.hk). Q. Wu is also with the Department of Electrical and Computer Engineering, Princeton University, Princeton, NJ 08544 USA.}
\thanks{T. Yang is with Meituan, Shenzhen, China. (e-mail: tianyu-yang@outlook.com)}
\thanks{J. Wan is with the School of Computer Science and Technology, Harbin Institute of Technology, Shenzhen 518066, China (e-mail:
jiawan1998@gmail.com).}
\thanks{Matias Di Martino is with the Department of Electrical and Computer Engineering, Duke University, Durham, NC 27708 USA.}
\thanks{Guillermo Sapiro is with the Department of Electrical and Computer Engineering, Princeton University, Princeton, NJ 08544 USA.}
}




\maketitle

\begin{abstract}
Tracking Any Point (TAP) has emerged as a fundamental tool for video understanding. Current approaches adapt Vision Foundation Models (VFMs) like DINOv2 via offline fine-tuning or test-time optimization. However, these VFMs rely on static image pre-training, which is inherently sub-optimal for capturing dense temporal correspondence in videos.
To address this, we propose Mask-to-Point (M2P) learning, which leverages rich video object segmentation (VOS) mask annotations to improve VFMs for dense point tracking.
Our M2P introduces three new mask-based constraints for weakly-supervised representation learning. 
First, we propose a local structure consistency loss, which leverages Procrustes analysis to model the cohesive motion of points lying within a local structure, achieving more reliable point-to-point matching learning.
Second, we propose a mask label consistency (MLC) loss, which enforces that sampled foreground points strictly match foreground regions across frames. The proposed MLC loss can be regarded as a regularization, which stabilizes training and prevents convergence to trivial solutions.
Finally, mask boundary constrain is applied to explicitly supervise boundary points.
We show that our weakly-supervised M2P models significantly outperform baseline VFMs with efficient training by using only 3.6K VOS training videos. Notably, M2P achieves 12.8\% and 14.6\% performance gains over DINOv2-B/14 and DINOv3-B/16 on the TAP-Vid-DAVIS benchmark, respectively. Moreover, the proposed M2P models are used as pre-trained backbones for both test-time optimized and offline fine-tuned TAP tasks, demonstrating its potential to serve as general pre-trained models for point tracking. Code will be made publicly available upon acceptance.
\end{abstract}


\section{Introduction}


Tracking Any Point (TAP) is a fundamental task that aims to accurately predict the 2D position and occlusion status of query points given in the first frame of a video. TAP is essential for wide-ranging applications, including robotics \cite{garcia2007evolution}, autonomous driving \cite{new_frame_cf,meta_graph,huang2018apolloscape,kristan2024second}, and view synthesis \cite{zhou2016view}. Inspired by deep learning, numerous trackers \cite{tumanyan2025dino,karaev2025cotracker,karaev2024cotracker3,li2025taptr} have achieved significant progress. Despite great success achieved by recent deep learning-based TAP approaches, they still suffer from severe tracking failures when faced with complex real-world challenges, such as fast motion, illumination variations, deformations, and long-term occlusions. The primary reasons for this can be attributed to: 1) the mainstream TAP paradigm involves training models in a fully-supervised (point-to-point) manner on densely-annotated synthetic datasets \cite{doersch2023tapir,locotrack}, which lack the diversity of motions and objects in natural videos; 2) these models are limited in their ability to aggregate information over long temporal context, making them hard to track points after significant occlusions.

  \begin{figure}
\begin{center}
   \includegraphics[width=0.85\linewidth]{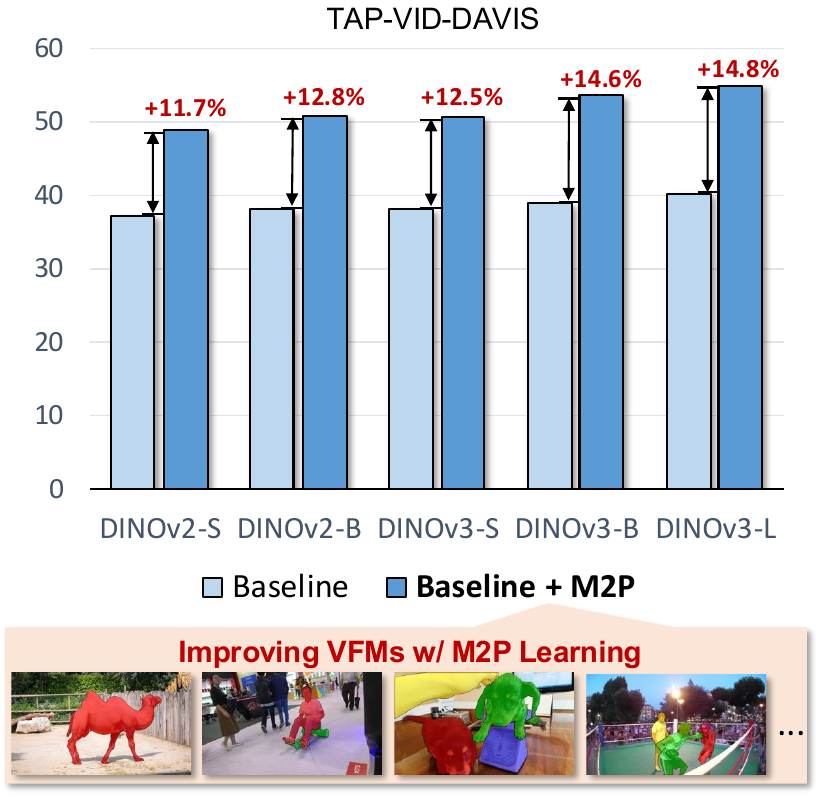} 
\end{center}
 \caption{Our proposed Mask-to-Point (M2P) learning improves visual foundation models (e.g., DINOs) on dense point tracking, while being trained on only 3.6K VOS videos.}
\label{page_fig}
\end{figure}


To ease the need of using large-scale point tracking datasets for training, current TAP approaches \cite{chrono,tumanyan2025dino} mainly leverage visual foundation models (VFMs) with strong semantic data-driven priors. For example, DINO-Tracker \cite{tumanyan2025dino} employs the DINOv2 \cite{oquab2023dinov2} backbone and learns a residual representation for test-time optimization. To facilitate learning on the synthetic Kubric dataset \cite{doersch2023tapir}, Chrono \cite{chrono} also uses the DINOv2 backbone for offline fine-tuning. Due to the strong semantic prior of VFMs, these trackers achieve state-of-the-art performance while maintaining simpler tracking frameworks. Despite great success of VFMs like DINOv2 in TAP, their performance remains limited, as these VFMs are trained on static images and thus inherently lack strong temporal matching ability.



Given an initial object mask on the first frame, video object segmentation (VOS) aims to recover the object's mask in subsequent video frames. 
At a high-level, VOS could be considered similar to the TAP task, but only focuses only on tracking the boundary points of the object. 
VOS has been extensively studied over the years, benefiting from large-scale real-world datasets with rich mask annotations (e.g., YouTube-VOS~\cite{youtubevos} and SA-V~\cite{ravi2024sam}). These datasets enable learning robust object- and boundary-level temporal correspondences. However, it remains unclear whether such object/boundary-level matching models can be effectively applied for fine-grained point matching. 
Moreover, despite rich temporal correspondences embedded in video mask annotations, existing TAP methods have not explored weakly-supervised learning of point tracking representations from them. Instead, most TAP methods still rely on pseudo or synthetic data for fully-supervised point matching learning.

In this work, we explore bridging the gap between TAP and VOS, and show that object mask annotations in VOS can be effectively used for weakly-supervised learning of point tracking representations (see Fig. \ref{page_fig}).
To achieve this, we propose Mask-to-Point (M2P) learning, which leverages the rich mask annotations in VOS datasets to enhance visual foundation models (VFMs) for dense point tracking.
M2P learns point-level correspondences under multiple mask-guided constraints. It first enforces local structure consistency by modeling the cohesive motion of points within a local structure region through Procrustes analysis, encouraging more reliable point-to-point matching. In addition, a mask label consistency loss ensures that sampled foreground points remain aligned with foreground regions across frames, acting as a regularizer that stabilizes training and avoids trivial collapse. Finally, M2P explicitly supervises boundary-proximal points using a mask boundary constraint to further improve performance.

Our M2P models significantly outperform current state-of-the-art VFMs, achieving 12.8\% and 14.6\% improvements over DINOv2-B/14 and DINOv3-B/16 on the TAP-VID DAVIS benchmark, respectively, while being trained on only 3.6K videos with mask annotations \cite{youtubevos,davis17}. We further use the learned M2P models as pre-trained backbones for downstream point tracking training, including test-time optimization (DINO-Tracker~\cite{tumanyan2025dino}) and offline fine-tuning (Chrono~\cite{chrono}) based frameworks. The pre-trained M2P models consistently yield superior results in both frameworks, demonstrating its potential as effective pre-trained backbones for TAP.

In summary, the main contributions of our work are:
\begin{compactitem}
  \item 
 We propose mask-to-point (M2P) weakly-supervised learning, which leverages VOS mask annotations to improve VFMs for dense point tracking. To the best of our knowledge, we are the first to investigate learning effective point tracking representation from VOS datasets.
 \item
We propose mask-guided learning objectives including a local structure consistency loss based on Procrustes analysis to model cohesive motion within local regions, and a mask label consistency loss that facilitates foreground-point matching across frames.
\item
We further apply a {mask boundary constraint} to explicitly supervise points near object boundaries, further improving tracking performance.
 \item
Our M2P models significantly improve over existing VFMs (e.g., DINOv2 and DINOv3) for TAP. Moreover, the learned M2P models can serve as general pre-trained backbones for downstream point tracking frameworks, including both test-time  optimization and offline fine-tuning based approaches.
 \end{compactitem}



\section{Related Work}

\noindent\textbf{Tracking Any Point (TAP).} 
The TAP task aims to localize query points and infer their occlusion states across video frames. Early progress includes PIPs~\cite{harley2022particle} for long-term tracking via iterative optimization and TAPNet~\cite{doersch2022tap}, which established the TAP-Vid benchmark. TAPIR~\cite{doersch2023tapir} improves TAPNet with a two-stage match–refine pipeline, while PointOdyssey~\cite{zheng2023pointodyssey} introduces a large-scale synthetic dataset and PIPs++ for stronger supervision. Transformer-based trackers further advance TAP, including Co-Tracker~\cite{karaev2025cotracker} for joint trajectory modeling, DOT~\cite{le2024dense} for unified point tracking and optical flow, TAPTR~\cite{li2025taptr} with a DETR-style architecture, and LocoTrack~\cite{locotrack} extending correlations to 4D volumes. More recent efforts exploit pseudo-labeled videos~\cite{doersch2024bootstap,karaev2024cotracker3}, high-resolution \cite{harley2025alltracker} and online memory \cite{dong2025online, aydemir2025track} designs, image-matching priors \cite{tan2025retracker}, multi-task learning \cite{badki2026l4p}, and sequential masked decoding \cite{zholus2025tapnext}.

Due to the difficulty in annotating precise point tracks on real videos,  the aforementioned approaches all employ the synthetic Kubric dataset \cite{doersch2023tapir} for training via fully-supervised point-to-point losses. To ease the need of training data, DINO-Tracker \cite{tumanyan2025dino} uses the foundation DINOv2 \cite{oquab2023dinov2} model as the main feature extractor and learns an additional residual adaptor for online adaptation. Chrono \cite{chrono} also employs DINOv2 as the backbone and adaptively learns the temporal adaptor using the synthetic data. Despite their success, VFMs like DINOv2 are still learned on static images, thus naturally lacking temporal matching ability and context. To overcome this issue, our work focuses on improving these VFMs for temporal matching in a weakly-supervised learning framework, using object masks rather than precise point data. We hope that our improved VFMs can benefit the TAP community.




\noindent\textbf{Video Object Segmentation (VOS).} 
Early VOS approaches mainly rely on online fine-tuning at test time to adapt to online target instances (e.g., OSVOS \cite{OSVOS}, OnAVIS \cite{OnaVOS}, MoNet \cite{MoNet}, MaskTrack \cite{MaskTrack} and PReMVOS \cite{PReMVOS}). While effective, these methods suffer from high latency and limited adaptation due to few online samples. Efficiency was improved using meta-learning and retrieval: OSMN \cite{OSMN} leverages a meta-network for fast adaptation with a single forward pass, while PML \cite{PML} treats VOS as pixel-wise retrieval in an embedding space. 
Subsequent works shifted towards matching frameworks. STM \cite{STM} pioneered memory-based matching, where past frames are used as external memory, which was then  enhanced through better integration (CFBI \cite{CFBI}, FEELVOS \cite{Feelvos}), multi-object association with transformers (AOT \cite{AOT}), efficient similarity metrics (STCN \cite{STCN}, SimVOS \cite{simvos}), large-scale generative pre-training \cite{dropmae,pul,wu2023dropmae} and sparse spatiotemporal modeling (SSTVOS \cite{SSTVOS}). Recently, XMEM \cite{XMEM} extended STM via an Atkinson–Shiffrin style memory, enabling robust long-term segmentation. SAM2 \cite{ravi2024sam} proposes the largest VOS dataset (SA-V), and learns a foundation segmentation model for both images and videos.

By considering the rich annotation sources in VOS, e.g., SA-V, DAVIS17 \cite{davis16,davis17} and Youtube-VOS \cite{youtubevos}, our work bridges the gap between TAP and VOS, and  explores the use of video mask annotations to improve VFMs (e.g., DINOv2 and DINOv3 \cite{dinov3}) for TAP. This new learning paradigm avoids using synthetic datasets or generating pseudo labels in videos, improving VFMs in a weakly-supervised way.





  \begin{figure*}
\begin{center}
   \includegraphics[width=1.0\linewidth]{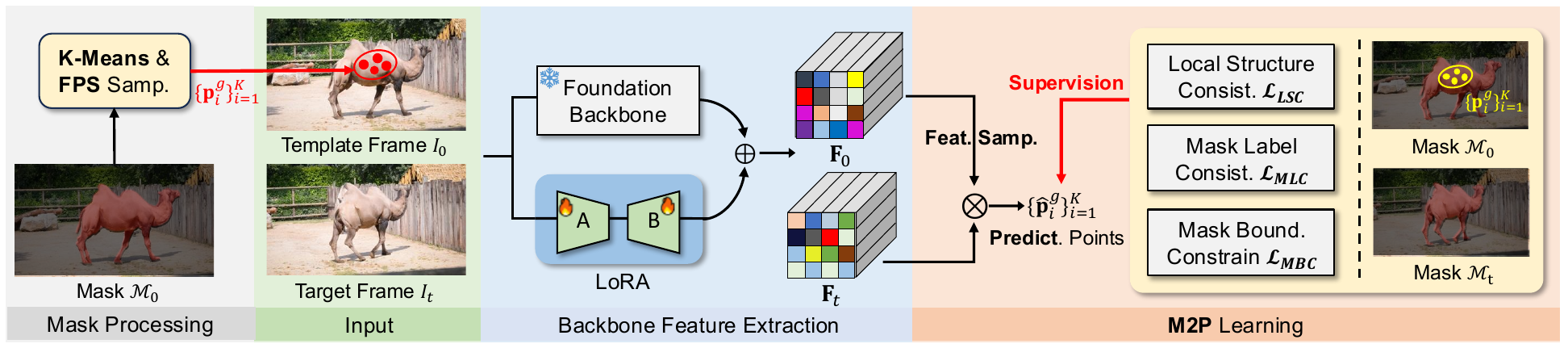}
\end{center}
\vspace{-0.2cm}
 \caption{Overall pipeline of the proposed Mask-to-Point (M2P) weakly-supervised learning, which enhances existing visual foundation models (VFMs) for dense point tracking by leveraging video object segmentation (VOS) datasets with mask annotations. Our M2P introduces three new mask-based constraints for representation learning, including \emph{local structure consistency}, \emph{mask label consistency} and \emph{mask boundary constrain}.}
\label{overall_pipe}
\end{figure*}

\section{Methodology}
In this section, we first revisit tracking any point (TAP) using VFMs and object-level temporal correspondence learning using mask annotations in VOS. Then we detail our proposed mask-to-point (M2P) weakly-supervised learning with VOS datasets to improve VFMs on TAP. Finally, we introduce the downstream applications of using our M2P pre-trained models for both online optimized and offline fine-tuned TAP tasks.

\subsection{Preliminary}

\noindent\textbf{TAP with VFMs.} Recent approaches employ pre-trained VFMs like DINOv2~\cite{oquab2023dinov2} to extract frame features $\bF \in \real^{T \times C \times H \times W} $ given an input video with $T$ frames. 
For a query point $\bp_{i}$, its feature $\bff_{i} \in \real^{1 \times C}$ is obtained by bilinear interpolation from the template frame feature $\bF_0$. A correlation map $\mathcal{C}_i \in \real^{H \times W}$ is then calculated as the cosine similarity between $\bff_{i}$ and the target frame feature $F_t$
\CUT{
\begin{equation}
\mathcal{C}_t = 
\frac{\bff_q \bF_{t}}
{\|\bff_q\|\,\|\bF_{t}\|},
\label{eq:corr}
\end{equation}
}
\begin{equation}
\mathcal{C}_i = \text{cos}(\bff_i, \bF_{t}),
\label{eq:corr}
\end{equation}
where $\text{cos}$($\cdot$) denotes cosine similarity computed along the channel dimension at each spatial location. 
The corresponding point location in target frame is then estimated over a refined map $H_{i} = \phi(\mathcal{C}_{i}) \in \real^{H\times W}$,
\begin{equation}
\tilde{\mathbf{p}}_i = 
\sum_{(x, y) \in \cal B} \sigma\big(H_i(x, y)\big) \cdot (x, y),
\label{eq:softargmax}
\end{equation}
where $\phi(\cdot)$ is a refiner head and $\sigma(\cdot)$ is the  softmax over all spatial positions within the correlation map. The above soft-argmax operation \cite{argsoftmax} computes a weighted spatial average of coordinates within a circular neighborhood ${\cal B}$ centered near the maximum-response region.

To enhance fine-grained point matching, existing methods \cite{chrono,tumanyan2025dino,aydemir2024can} typically employ DINOv2 \cite{oquab2023dinov2} to extract dense feature maps $\bF$, and further train domain-specific adapters to incorporate complementary information. Despite their success,  DINOv2-like VFMs are learned from static images, which is inherently suboptimal for capturing dense temporal correspondence in videos.

\CUT{
 the spatial softmax $\sigma(\cdot)$ is defined with a temperature parameter $\tau$ as:
\begin{equation}
\sigma\!\big(C_t(x, y)\big) =
\frac{\exp\!\big(\tau\,C_t(x, y)\big)}
{\sum_{(x', y')} \exp\!\big(\tau\,C_t(x', y')\big)}.
\label{eq:softmax}
\end{equation}

This non-learnable yet differentiable soft-argmax formulation provides per-frame 
position estimates directly from dense feature correlations, and can be further 
refined by masking a local neighborhood around the correlation peak. However, existing approaches mainly leverage image-based visual foundation models like DINOv2, which is not optimal for point tracking in videos, where the main focus is temporal matching.
}

\noindent\textbf{Temporal Correspondence Learning in VOS.} Existing VOS approaches \cite{simvos,ravi2024sam} leverage object-level video mask annotations for temporal correspondence learning. Given a template frame $I_{0}$ with its object annotation mask $\mathcal{M}_{0}$, VOS models aim to predict the corresponding mask $\mathcal{M}_{t}$ in the $t$-th frame. The training procedure is formulated as
\begin{equation}
\mathcal{L}_{vos} = \Phi\!\bigl( V(\mathcal{M}_{0},\, I_{0},\, I_{t}),\, \mathcal{M}_{t} \bigr),
\label{eq:vos_learn}
\end{equation}
where $V(\cdot)$ is the VOS model and $\Phi(\cdot)$ is the cross-entropy loss commonly used in VOS. Although the current VOS models can effectively perform object-level temporal matching in complex scenarios,  it remains unclear whether such object-level matching models can benefit fine-grained point matching.

We confirm this via experiments presented  in Table \ref{tab:zero_tapvid_strided},  where VOS backbones, e.g., SAM2 \cite{ravi2024sam} and SimVOS \cite{simvos}, fail to perform fine-grained point matching across video frames via (\ref{eq:corr}), which indicates that the models learned with Eq.~\ref{eq:vos_learn} mainly capture object-level temporal correspondences. 
Considering the rich temporal correspondences embedded in video mask annotations, we propose mask-to-point (M2P) learning, which can effectively learn point tracking representations from VOS datasets.



\subsection{Mask-to-Point Weakly-Supervised Learning}
Our goal is to boost the fine-grained point tracking ability of VFMs by leveraging richly annotated mask data in VOS. 
The overall pipeline of M2P learning is presented in Fig.~\ref{overall_pipe}.

\noindent\textbf{Query Point Sampling.} Given a template frame $I_{0}$ and a target frame $I_{t}$ with their corresponding mask annotations $\mathcal{M}_{0}$ and $\mathcal{M}_{t}$, we first perform query point sampling on the template frame. To obtain diverse and representative query points, we first perform K-means clustering on the pixels within the mask region $\mathcal{M}_{0}$ based on their spatial coordinates, partitioning them into $G$ spatial groups. Within each group, we then apply farthest point sampling (FPS)~\cite{qi2017pointnet++} to select $K$ spatially diverse points within each group. Consequently, for the $g$-th group, we obtain a query point set $\bQ_{g} = \{\bp_{i}^{g}\}_{i=1}^{K}$, and the overall query set is defined as $\bQ = \bigcup_{g=1}^{G} \bQ_{g}$.

\noindent\textbf{Local Structure Consistency (LSC).} The sampled points within the same group $\bQ_{g}$ are spatially close and tend to lie within the same local structural region. As the number of groups $G$ increases, points in each group become more concentrated, thus exhibiting more similar motion patterns across frames. 
Leveraging this property, we assume that all points within a local structure group undergo a locally consistent motion that can be approximated by a local similarity transformation (i.e., rigid transformation with scale). 
Based on this assumption, we estimate the group-wise motion from a few reliable correspondences and propagate it to the remaining points for supervision.


To get reliable point correspondences for motion transform estimation, we first employ an off-the-shelf VFM to predict point-wise correspondences between the query points and the target frame via (\ref{eq:corr}). The matched points in the target frame are obtained by applying the argmax operation on $\mathcal{C}_{t}$. 
We rank all query points in each group by their matched scores and select the top-$K_{e}$ most confident pairs as reliable correspondences, 
\begin{equation}
\mathcal{P}_{g} = \{(\bp^{g}_{i}, \tilde{\bp}^{g}_{i})\}_{i=1}^{K_{e}}, \quad K_{e}<K,
\label{point_set}
\end{equation}
where $\tilde{\bp}^{g}_{i}$ denotes the predicted correspondence of $\bp^{g}_{i}$.

 \begin{figure}
\begin{center}
   \includegraphics[width=1.0\linewidth]{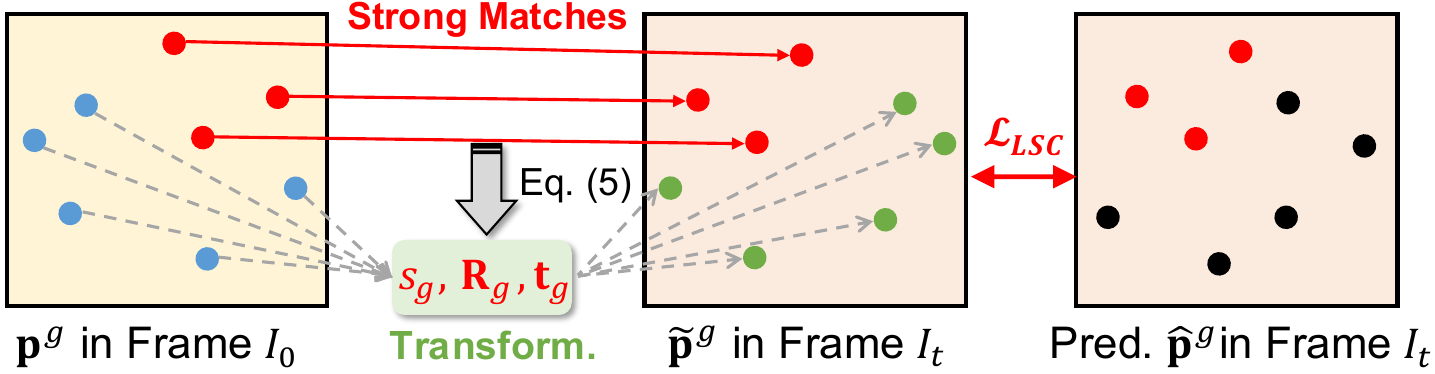} 
\end{center}
\vspace{-0.2cm}
 \caption{Illustration of our proposed local structure consistency loss $\mathcal{L}_{LSC}$. The top-$K_e$ strong matches are found between query points $\bp^g$ and predicted correspondences,  $\{\tilde{\bp}^g\}_{i=1}^{K_e}$. These strongly matched points are used to estimate a simalarity transformation, which is then applied to the other points to obtain remaining pseudo-labels $\tilde{\bp}^g_i$, $i>K_e$, for supervision.}
\label{illus_lsc}
\end{figure}

Given the top-$K_{e}$ confident correspondences $\mathcal{P}_{g}$, we  estimate a similarity transformation 
parameterized by rotation $\mathbf{R}_{g}\in\real^{2\times2}$, translation $\mathbf{t}_{g}\in\real^{2}$, 
and isotropic scale $s_{g}\in\real$ that aligns the template points to their predicted target locations. 
This transformation can be estimated via Procrustes analysis \cite{schonemann1966generalized} (please refer to the Supplementary for more details), which finds the optimal similarity transform minimizing the squared alignment error given by
\begin{equation}
\min_{s_{g},\,\mathbf{R}_{g},\,\mathbf{t}_{g}} 
\sum_{i=1}^{K_{e}}
\left\|
\tilde{\bp}^{g}_{i} - 
\left(s_{g}\mathbf{R}_{g}\bp^{g}_{i} + \mathbf{t}_{g}\right)
\right\|^{2}.
\label{eq:procrustes}
\end{equation} 

\noindent The estimated transformation is then applied to the points in $\bQ_{g}$ to obtain their matched positions in the target frame:
\begin{equation}
\tilde{\bp}^{g}_{j} = s_{g}\mathbf{R}_{g}\bp^{g}_{j} + \mathbf{t}_{g}, 
\quad j = K_e+1, \dots, K.
\end{equation} 

\noindent Following \cite{tumanyan2025dino,chrono,karaev2025cotracker} we supervise the predicted correspondences using a robust Huber loss between the transformed points and their predicted target locations,
\begin{equation}
\mathcal{L}_{\text{LSC}} 
= 
\frac{1}{G K}
\sum_{g=1}^{G}
\sum_{i=1}^{K}
\text{Huber}\!\left(
\tilde{\bp}^{g}_{i}, 
\hat{\bp}^{g}_{i}
\right),
\label{eq:lsc_loss}
\end{equation}
where $\tilde{\bp}^{g}_{i}$ is the transformed position of the $i$-th point in the group $g$, and $\hat{\bp}^{g}_{i}$ denotes the corresponding location of $\bp_{i}^{g}$ in the target frame obtained via (\ref{eq:softargmax}).

We detail the learning steps of LSC in Fig.~\ref{illus_lsc}. In our implementation, we use the top $K_{e}=3$ most confident matched points to estimate the similarity transformation in (\ref{eq:procrustes}), since 
three unique matched points are sufficient to determine a 2D similarity transformation including rotation, translation and scale. 
In addition, using only a few highly reliable correspondences also alleviate the impact of noisy predictions, which could otherwise lead to inaccurate motion estimation.

\noindent\textbf{Mask Label Consistency.} Since we sample query points within the foreground mask region $\mathcal{M}_{0}$ of the template frame $I_{0}$, their corresponding points in the target frame $I_{t}$ should also lie within the foreground region. 
This observation allows us to impose a mask label consistency supervision \yty{on the predicted correspondence. 
Specifically, we define a confidence score $\mathcal{S}_{i}^g$ to  represents the probability that the predicted correspondence of $\bp_{i}^g$ lies inside the target foreground region.}
\begin{equation}
\mathcal{S}_{i}^g = \sum_{h=1}^{H}\sum_{w=1}^{W} \left(\sigma(H_{i}^{g}) \odot \mathcal{M}_t\right)_{h,w}
\label{eq:softmask}
\end{equation}
\yty{where $\sigma(H_{i}^{g}) \in \mathbb{R}^{H\times W}$ is the correlation map defined in ~(\ref{eq:softargmax}) and $\mathcal{M}_t\in\mathbb{R}^{H\times W}$ denotes the binary mask annotation of target frame (i.e., 1 for foreground, 0 for background).} 

\CUT{. Specifically, given a query point $\bp_{q}$, we first obtain its softmax correlation map $\sigma(H_{t}^{q}) \in \mathbb{R}^{1\times HW}$ as defined in (\ref{eq:softargmax}). 
We then compute its soft foreground confidence by taking a weighted sum over the target-frame mask, where the weights are given by the softmax correlation map,
\begin{equation}
\mathcal{S}_{q} = \sigma(H_{t}^{q})\,{\mathcal{M}_{t}}^{\top},
\label{eq:softmask}
\end{equation}
where $\mathcal{M}_{t}\in\mathbb{R}^{1\times HW}$ denotes the binary mask annotation (i.e., 1 for foreground, 0 for background). Fig.~\ref{illus_mlc} shows the calculation of $S_{q}$. Intuitively, $\mathcal{S}_{q}$ represents the probability that the predicted correspondence of $\bp_{q}$ lies inside the target foreground region.}

 \begin{figure}
\begin{center}\includegraphics[width=1.0\linewidth]{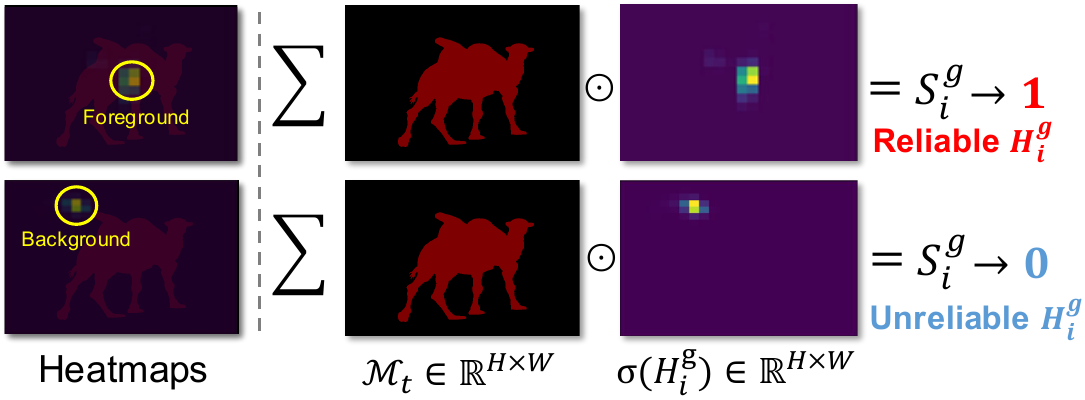} 
\end{center}
\vspace{-0.2cm}
 \caption{Illustration of the confidence score $\mathcal{S}_i^g$ in our mask label consistency constraint. Zoom in for clearer visualization.}
\label{illus_mlc}
\end{figure}

In practice, query points located near mask boundaries may produce correlation maps that slightly extend beyond the foreground region. As a result, even correct correspondences can yield soft labels $\mathcal{S}_{i}^g$ lower than~1 due to partial spatial overlap. To handle these cases, we introduce a tolerance threshold~$\tau=0.5$ and apply the loss only when $\mathcal{S}_{i}^g\le\tau$,
\begin{equation}
\mathcal{L}_{\text{MLC}}
=
\frac{1}{G K}
\sum_{g=1}^{G}
\sum_{i=1}^{K}
\begin{cases}
0, & \text{if } \mathcal{S}_{i}^g > \tau,\\[3pt]
-\log(\mathcal{S}_{i}^g + \epsilon), & \text{otherwise.}
\end{cases}
\label{eq:l_mlc}
\end{equation}

Empirically, we find that the proposed mask label consistency loss $\mathcal{L}_{\text{MLC}}$ acts as an effective regularization term during training. As seen in Fig.~\ref{abl_figs}(a), it stabilizes the optimization process and prevents the model from collapsing into trivial correspondence patterns. 
By enforcing foreground-to-foreground consistency, $\mathcal{L}_{\text{MLC}}$ encourages the model to learn meaningful and object-driven coherent matching behavior, which further enhances the overall point matching performance.

\noindent\textbf{Mask Boundary Constraint.} We observe that query points located on or near the mask boundary tend to have their corresponding points positioned similarly close to the boundary, since these pixels physically lie around object edges.
Based on this observation, we introduce a \emph{mask boundary constraint} (MBC) loss that explicitly enforces consistency in the relative distance to the mask boundary across frames.
This loss also encourages the predicted points to span the entire object, even when the object size changes.

\CUT{
 \begin{figure}
\begin{center}
   \includegraphics[width=0.85\linewidth]{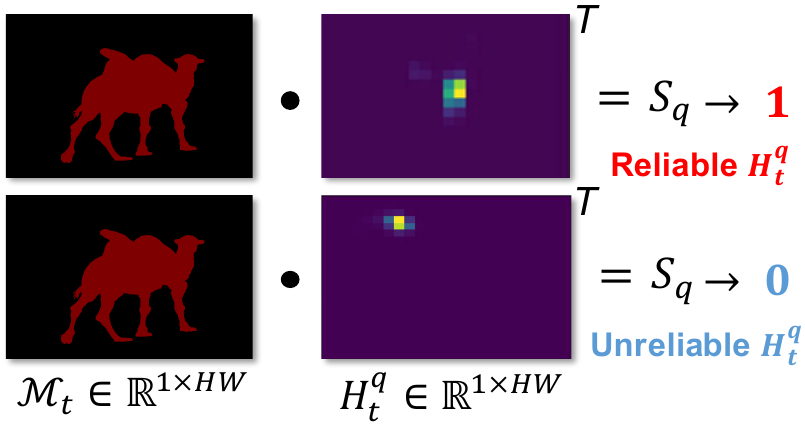} 
\end{center}
\vspace{-0.6cm}
 \caption{Illustration of our mask label consistency constrain. Zoom-in for better visualization.}
\label{illus_mlc}
\end{figure}
}

Specifically, for the $g$-th group of sampled points, we compute the normalized distance from each point to the mask boundary in both the template and target frames, 
denoted as $\tilde{d}^{0}_{g,i}$ and $\tilde{d}^{t}_{g,i}$, respectively,
\begin{equation}
\tilde{d}^{0}_{g,i} = 
\tfrac{d^{0}_{g,i}}{\sum_{i=1}^{K} d^{0}_{g,i} + \epsilon}, 
\quad
\tilde{d}^{t}_{g,i} = 
\tfrac{d^{t}_{g,i}}{\sum_{i=1}^{K} d^{t}_{g,i} + \epsilon},
\end{equation}
where $d^{0}_{g,i}$ and $d^{t}_{g,i}$ denote the Euclidean distances of the $i$-th point to the nearest foreground mask boundary pixel in the template and target masks, respectively. $\epsilon$ is a small constant added for numerical stability. 
To emphasize points closer to the boundary, we introduce a distance weighting function
\begin{equation}
w_{g,i} = \exp(-d^{0}_{g,i}).
\end{equation}
The final MBC loss \CUT{for the $g$-th group }is then defined as
\begin{equation}
\mathcal{L}_{MBC} = \frac{1}{G}\sum_{g=1}^{G} 
\tfrac{
\sum_{i=1}^{K} 
w_{g,i} \cdot 
\left| 
\tilde{d}^{0}_{g,i} - 
\tilde{d}^{t}_{g,i} 
\right|
}{
\sum_{i=1}^{K} w_{g,i} + \epsilon
},
\label{eq:l_mbc}
\end{equation}
which explicitly regularizes points located near object boundaries to maintain consistent relative distances to the contour across frames.

\begin{algorithm}[t]
\caption{Mask-to-Point (M2P) Learning}
\label{alg:m2p}
\begin{algorithmic}[1]
\STATE \textbf{Input:} Frame pair $(I_0, I_t)$ with masks $(\mathcal{M}_0, \mathcal{M}_t)$.
\STATE \textbf{Parameters:} VFM backbone, groups $G$, points per group $K$, weights $\{\lambda_1, \lambda_2, \lambda_3\}$.

\STATE // \textit{Query Point Sampling}
\STATE Cluster $\mathcal{M}_0$ into $G$ groups, sample $K$ points per group via FPS.
\STATE Obtain query set: $\bQ = \bigcup_{g=1}^G \bQ_g$, where $\bQ_g = \{\bp_i^g\}_{i=1}^K$.

\STATE // \textit{Point Matching}
\STATE Extract features: $\bF_0, \bF_t$ from $I_0, I_t$ via VFM backbone.
\FOR{each $\bp_i^g \in \bQ$}
    \STATE Compute $\mathcal{C}_i = \text{cos}(\bff_i, \bF_t)$ and predict $\hat{\bp}_i^g$ (Eqs.~\ref{eq:corr}--\ref{eq:softargmax}).
\ENDFOR

\STATE // \textit{Loss Computation}
\STATE Compute $\mathcal{L}_{\text{LSC}}$ via Procrustes transform (Eq.~\ref{eq:lsc_loss}).
\STATE Compute $\mathcal{L}_{\text{MLC}}$ from $\mathcal{M}_t$ (Eq.~\ref{eq:l_mlc}).
\STATE Compute $\mathcal{L}_{\text{MBC}}$ from boundary distances (Eq.~\ref{eq:l_mbc}).

\STATE // \textit{Optimization}
\STATE $\mathcal{L}_{\text{total}} \leftarrow \lambda_1 \mathcal{L}_{\text{LSC}} + \lambda_2 \mathcal{L}_{\text{MLC}} + \lambda_3 \mathcal{L}_{\text{MBC}}$.
\STATE Update VFM parameters via backpropagation on $\mathcal{L}_{\text{total}}$.
\end{algorithmic}
\end{algorithm}

\noindent\textbf{Joint Training.} Our M2P uses the proposed objectives to jointly  learn spatially coherent (via local structure consistency and boundary constraint) and semantically consistent (via foreground consistency) point correspondences. The overall training objective is formulated as a weighted combination of the three losses,
\begin{equation}
\mathcal{L}_{\text{total}} = 
\lambda_{1}\mathcal{L}_{\text{LSC}} + 
\lambda_{2}\mathcal{L}_{\text{MLC}} + 
\lambda_{3}\mathcal{L}_{\text{MBC}},
\label{eq:joint_loss}
\end{equation}
where the hyperparameters $\lambda_{1}$, $\lambda_{2}$, and $\lambda_{3}$ control the relative importance of each term. Algorithm~\ref{alg:m2p} summarizes the overall procedure of our proposed Mask-to-Point (M2P) learning.

%
%
%

\subsection{Mask-to-Point Learning Algorithm}

Algorithm~\ref{alg:m2p} summarizes the overall procedure of our proposed Mask-to-Point (M2P) learning framework.
Given a video frame pair with mask annotations, M2P learns point-level correspondences through three mask-guided constraints.

 \begin{figure*}
\begin{center}
   \includegraphics[width=0.9\linewidth]{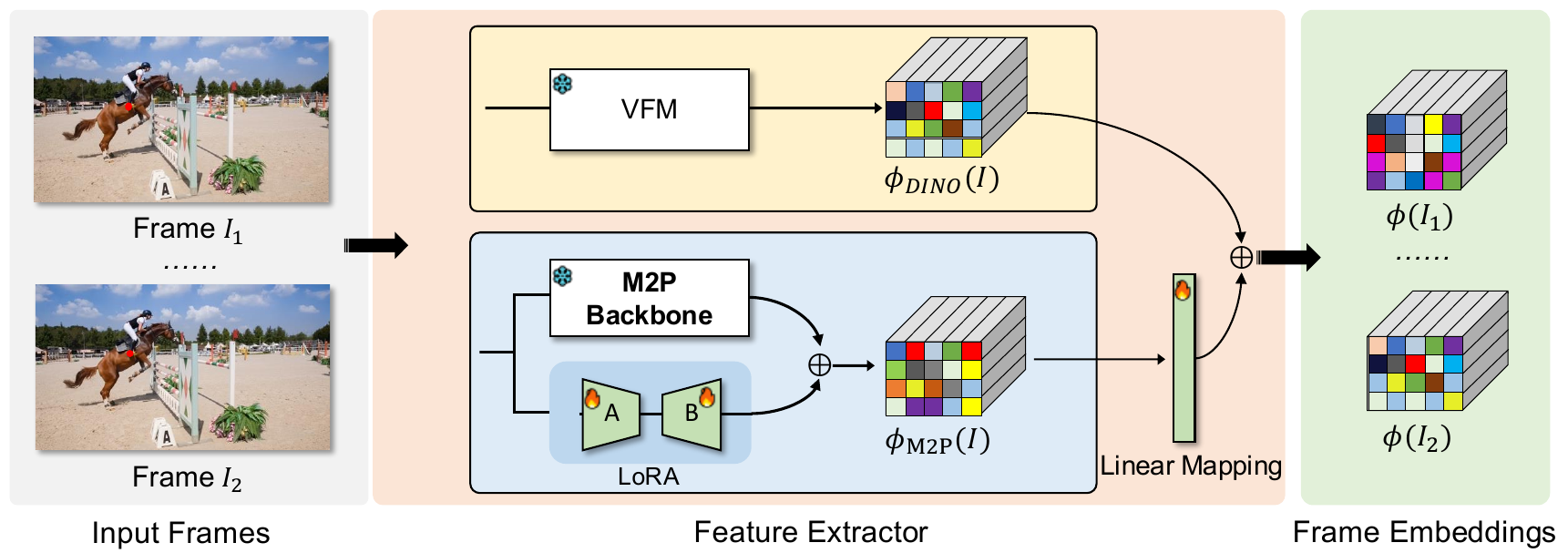} 
\end{center}
\vspace{-0.2cm}
 \caption{An overall pipeline of the proposed M2P-Tracker for test-time optimization TAP. The proposed M2P-Tracker uses our M2Pv3-S/16 backbone for feature extraction, providing strong temporal prior for more effective online adaptation.}
\label{m2p_tracker_pipeline}
\end{figure*}

\noindent\textbf{Architecture.} We employ VFMs such as DINOv2 and DINOv3 as our backbones. Following~\cite{tumanyan2025dino,aydemir2024can}, we implement $\phi(\cdot)$ as a lightweight point module composed of two convolutional layers. For efficient pre-training, we adopt Low-Rank Adaptation (LoRA) \cite{hu2022lora} in each block of the VFM backbones, enabling parameter-efficient fine-tuning without updating the entire backbones (see Tables \ref{tab:zero_tapvid_strided} and \ref{tab:lora_ablation}).

\CUT{
\noindent\textbf{Mask Boundary Constrain.} For the sampled points close to the mask boundary, we suppose that their corresponding points are also close to the boundary, especially considering that most of them are physically located near the boundary. To further supervise these points, we further propose a mask boundary constrain (MBC) loss, which uses point-to-boundary distance-based supervision:

As can be seen, points closer to the boundary have larger importance for supervision, and norm point-to-boundary distance makes sure that these boundary points have consistent corresponding points.
}

\subsection{Downstream Applications with M2P Models}
After pre-training the backbones on VOS datasets using the loss function in (\ref{eq:joint_loss}), 
our M2P models 
can be further leveraged for various downstream TAP applications.

\noindent\textbf{Weakly-Supervised Point Tracking.}
Following \cite{aydemir2024can}, without applying any synthetic data fine-tuning or online optimization, our pre-trained M2P models are employed for direct point tracking as formulated in (\ref{eq:softargmax}). Since our M2P models do not use any domain-specific annotated TAP data for learning, this setting 
is Weakly-Supervised (WS) Point Tracking.

\noindent\textbf{Offline Fine-tuning TAP.} Due to the large annotation cost, current approaches \cite{chrono,locotrack,karaev2025cotracker,zholus2025tapnext} mainly employ the synthetic Kubric dataset \cite{doersch2023tapir} for point temporal correspondence learning. To verify the effectiveness of our M2P models in synthetic data fine-tuning, we replace the ViT \cite{ViT} backbone in \cite{chrono} with our M2P model for fine-tuning, and name this new variant as \textbf{M2P-Chrono}.

\noindent\textbf{Test-Time Optimization TAP.} 
DINO-Tracker \cite{tumanyan2025dino} is a representative test-time optimized TAP tracker, consisting of a motion estimator (Delta-DINO) and a VFM.
In our M2P-Tracker (see Fig.~\ref{m2p_tracker_pipeline}), we replace the Delta-DINO module with our M2Pv3-S/16 backbone for test-time optimization. 
During optimization, our M2Pv3-S/16 serves as the motion adapter, providing strong temporal priors that are complementary to the rich semantic representations from the DINOv2-L/14 VFM.


Specifically, given an input image, M2Pv3-S/16 is first used to extract image features. To align the feature dimension with that of DINOv2-L/14, we introduce a single $1\times1$ convolutional layer for channel mapping. For the point-prediction head, we adopt the same lightweight head used in M2Pv3-S/16, consisting of two convolutional layers with a hidden size of 16. Following \cite{tumanyan2025dino}, we apply the same training objectives and employ LoRA (rank = 8, applied to QKV) for test-time optimization. The LoRA training is applied to the last six self-attention layers of our M2Pv3-S/16 backbone. 
Benefiting from the strong temporal prior of M2P, our M2P-Tracker requires 15× fewer learnable parameters and only 25\% of the training iterations compared with the original DINO-Tracker, while achieving superior overall point accuracy and visibility prediction.

\vspace{-0.25cm}
\section{Experiments}

\noindent\textbf{Implementation Details.} We use VFMs including DINOv2 \cite{oquab2023dinov2} and DINOv3 \cite{dinov3} as our backbones, and rely on each chosen backbone to generate reliable point correspondences $\{\tilde{\bp}^{g}_{i}\}_{i=1}^{K_{e}}$. The LoRA \cite{hu2022lora} adapters are applied to \emph{atten}, \emph{proj} and \emph{fc} of each transformer block. 
For the offline fine-tuning and test-time optimization based TAP, we replace the backbones in Chrono \cite{chrono} and Dino-Tracker \cite{tumanyan2025dino} with our M2Pv3-S/14 backbone for high efficiency. 
For the point sampling, we both set the group $G$ and point number $K$ to 12. $\lambda_{1}$, $\lambda_{2}$ and $\lambda_{3}$ are empirically set to 0.02, 0.5, 10.0, respectively. 

 \begin{table}[t]
\vspace{-0.3cm}
  \newcommand{\tabincell}[2]
    \centering
 \small
    \begin{tabular}{c|c}
    \Xhline{\arrayrulewidth}
    \multicolumn{1}{c|}{Config}&  \multicolumn{1}{c}{Value}\cr
       \Xhline{\arrayrulewidth}  
    optimizer           &AdamW \cite{Adam}   \cr  
   base learning rate             & 2.0e-5      \cr 
     weight decay             &1.0e-7    \cr  
     batch size              &16       \cr 
     training iteration           & 150,000    \cr  
     lora rank           & 64    \cr  
     lora target modules           & QKV, Proj, FC    \cr  
     lora dropout ratio        & 0.1    \cr  
     learning rate (lr) schedule            &torch.MultiStepLR      \cr  
     lr decay step            & 110,000     \cr
     lr decay factor            & 0.1     \cr  
     data augmentation        & RandomAffine, ColorJitter    \cr  
   \Xhline{\arrayrulewidth}  
   \end{tabular}
   \centering
  \caption{The pre-training setting for our M2P learning.}  
  \label{pretraining_para}
\end{table}

\noindent\textbf{Evaluation Datasets.} 
We evaluate our method on the TAP-Vid benchmark~\cite{doersch2022tap}, including real-world collected videos used for dense point tracking. These evaluation videos provide accurate human-annotated point trajectories. TAP-Vid is mainly composed of TAP-Vid-Kinetics~\cite{k400} and TAP-Vid-DAVIS~\cite{davis17}. Specifically, TAP-Vid-Kinetics includes 1,189 YouTube videos that present substantial challenges such as heavy motion blur and sudden scene changes. TAP-Vid-DAVIS contains 30 real videos characterized by complex object motions and large scale\&appearance variations. Following DINO-Tracker \cite{tumanyan2025dino}, we evaluate the proposed M2P-Tracker on TAP-VID-DAVIS 480, which provides a 480–resolution variant of TAP-Vid-DAVIS.

\noindent\textbf{Pre-Training Datasets.} For our M2P training, we use the training splits of two video object segmentation datasets including YouTube-VOS 2019 (denoted as Y) \cite{youtubevos} and DAVIS-2017 (D) \cite{davis17}. Importantly, the DAVIS-2017 training set does not overlap with the TAP-VID-DAVIS test split, and the overall training data (Y+D) contains 3.4K videos. We demonstrate that M2P can be effectively learned from these videos, and the training is performed with a batch size 16 and LR=2e-5, which takes about one day on two RTX 4090 GPUs. Pre-training is conducted at a fixed feature map resolution of $24\times24$. The input image resolution is adjusted according to the backbone patch size.

\noindent\textbf{Downstream Training Details.} The training details for our M2P pre-training and offline TAP fine-tuning (M2P-Chrono) are shown in Tables~\ref{pretraining_para} and \ref{kubric_training}, respectively. All backbones used in M2P pre-training are trained under the same configuration for fair comparison. For the test-time optimized M2P-Tracker, we follow the same training setup as DINO-Tracker, but reduce the number of training iterations from 10,000 to 2,500, thanks to the strong temporal prior provided by our M2P models.

 \begin{table}[t]
\vspace{-0.3cm}
  \newcommand{\tabincell}[2]
    \centering
 \small
    \begin{tabular}{c|c}
    \Xhline{\arrayrulewidth}
    \multicolumn{1}{c|}{Config}&  \multicolumn{1}{c}{Value}\cr
       \Xhline{\arrayrulewidth}  
    optimizer           &AdamW \cite{Adam}   \cr  
   base learning rate             & 1.5e-4      \cr 
     weight decay             &1.0e-4    \cr  
     batch size              &4       \cr 
     training iteration           & 100,000    \cr 
     warm iteration 		& 300    \cr 
     learning rate (lr) schedule            &OneCycleLR      \cr  
   \Xhline{\arrayrulewidth}  
   \end{tabular}
   \centering
  \caption{The training setup for offline TAP fine-tuning on the synthetic Kubric \cite{doersch2023tapir} dataset..}  
  \label{kubric_training}
\end{table}

\noindent\textbf{Evaluation Metrics.} We evaluate tracking results using position accuracy at multiple thresholds and the average score $\delta^{x}_{avg}$. Specifically, $\delta^{0}$ to  $\delta^{4}$ measure the percentage of visible ground-truth points whose predictions fall within 1, 2, 4, 8, and 16 pixels, respectively, and $\delta^{x}_{avg}$ is the mean accuracy. Following prior work, evaluations are conducted under the strided-query and first-query settings. For test-time optimization TAP, we additionally report Occlusion Accuracy (OA) for visibility evaluation and Average Jaccard (AJ) for joint position-and-visibility performance. We also adopt $\delta^{seg}$ \cite{tumanyan2025dino} to assess keypoint accuracy.

\begin{table*}[t]
\centering
\small
\setlength{\tabcolsep}{5.0pt}
\renewcommand{\arraystretch}{1.2}
\begin{tabular}{l c c c c c c c}
\toprule
\textbf{Model} & \textbf{Setup} & \textbf{Size} & \textbf{\#L.P.} & \textbf{Data} & \textbf{TAP-VID-DAVIS} & \textbf{TAP-VID-Kinetics} & \textbf{RGB-Stacking} \\
\hline
MAE \cite{mae} & ZS & 32$\times$32 & - & S & 23.5 & 27.6 & 43.2\\
DeiT \cite{deit} & ZS & 32$\times$32 &- & S & 24.0 & 22.4 & 23.3 \\
CLIP \cite{clip} & ZS & 32$\times$32 & - & S & 25.4 & 25.0 & 33.8 \\
DINO \cite{dino} & ZS & 32$\times$32 & - & S & 34.5 & 34.4 & 39.3 \\
DINOv2-\textbf{S}/14 \cite{oquab2023dinov2} &ZS & 32$\times$32 & - & S & 37.1 & 33.4 & 36.9 \\
DINOv2-\textbf{B}/14-Reg \cite{oquab2023dinov2} &ZS & 32$\times$32 & - & S & 37.4 & 33.6 & 35.9 \\
DINOv2-\textbf{B}/14 \cite{oquab2023dinov2} &ZS & 32$\times$32 & - & S & 38.0 & 34.5 & 37.8 \\
DINOv3-\textbf{S}/16 \cite{dinov3} &ZS & 32$\times$32 & - & S & 38.0 & 36.7 & 43.3 \\
DINOv3-\textbf{B}/16 \cite{dinov3} &ZS& 32$\times$32 & - & S & 38.9 & 37.4 & 44.7 \\
SD \cite{stablediffusion} & ZS & 32$\times$32 &- & S & 33.9 & 37.2 & 46.2 \\
\hline
SMTC \cite{qian2023semantics} & ZS & 30$\times$45 &- & S & 33.9 & 36.1 & - \\
DIFT (SD1.5)  \cite{tang2023emergent} &ZS& 30$\times$45 & - & S & 38.2 & 41.9 & - \\
DIFT (SD2.1)  \cite{tang2023emergent} &ZS& 30$\times$45 & - & S & 39.7 & 42.9 & - \\
VFS (ResNet-50) \cite{vfs} & ZS & 30$\times$45 &- & V & 38.4 & 42.4 & - \\
CRW (ResNet-18) \cite{crw} & ZS & 30$\times$45 &- & V & 35.9 & 40.9 & - \\
DiffTrack (HunyuanVideo) \cite{difftrack} &ZS& 30$\times$45 & - & V & 44.1 & 45.5 & - \\
DiffTrack (CogVideoX-2B) \cite{difftrack} &ZS& 30$\times$45 & - & V & 46.3 & 46.3 & - \\
DiffTrack (CogVideoX-5B) \cite{difftrack} &ZS& 30$\times$45 & - & V & 46.9 & 49.2 & - \\
\hline
SAM  \cite{sam}& WS & 32$\times$32 & - & S & 29.5 & 31.4 & 44.7 \\
SimVOS-\textbf{B}/16 \cite{simvos} & WS & 32$\times$32 & - & D + Y (3.4k) & 30.1 & 32.0 & 43.2 \\
SAM2-\textbf{B}-Plus \cite{ravi2024sam} & WS & 32$\times$32 & - & SA-V (50k) & 32.4 & -  & 43.9 \\
SAM2-\textbf{L} \cite{ravi2024sam} & WS & 32$\times$32 & - & SA-V (50k) & 34.4 & - & - \\
\rowcolor{gray!6}
\textbf{M2Pv2}-\textbf{S}/14-R@64 & WS & 32$\times$32 & 4.7M & D + Y (3.4k) & {48.8} (+\textbf{11.7\%}) & 45.5 (+\textbf{12.1\%}) & 51.7 (+\textbf{14.8\%})  \\
\rowcolor{gray!6}
\textbf{M2Pv2}-\textbf{B}/14-R@64 & WS & 32$\times$32 & 9.5M & D + Y (3.4k) & {50.8} (+\textbf{12.8\%}) & 46.4 (+\textbf{11.9\%}) & 51.5 (+\textbf{13.7\%}) \\
\rowcolor{gray!6}
\textbf{M2Pv3}-\textbf{S}/16-R@64 & WS & 32$\times$32 & 4.7M & D + Y (3.4k) & {50.5} (+\textbf{12.5\%}) & 48.4 (+\textbf{11.7\%}) & 56.5 (+\textbf{13.2\%}) \\
\rowcolor{gray!6}
\textbf{M2Pv3}-\textbf{B}/16-R@64 & WS & 32$\times$32 & 9.5M & D + Y (3.4k) & \textbf{53.5} (+\textbf{14.6\%}) & \textbf{49.4} (+\textbf{12.0\%}) & \textbf{58.0} (+\textbf{13.3\%})  \\
\hline
\textcolor{gray}{Chrono-\textbf{S}/16} \cite{chrono} & \textcolor{gray}{FS} & 32$\times$32 & \textcolor{gray}{16.2M} & \textcolor{gray}{Kubric (11k)} & \textcolor{gray}{48.0} & \textcolor{gray}{{49.3}} & \textcolor{gray}{{67.0}} \\
\bottomrule
\end{tabular}
\caption{\textbf{Comparison on TAP-VID-DAVIS \cite{davis17}, TAP-VID-Kinetics \cite{k400} and RGB-Stacking \cite{doersch2022tap} under the queried-first mode}. Models are commonly evaluated at a final feature map resolution of  $32 \times 32$ or $30 \times 45$, following the original evaluation in image-based and video-based models. ZS, WS and FS denote {zero-shot}, {weakly-supervised} and {fully-supervised} learning settings, respectively. D \cite{davis17}, Y \cite{youtubevos}, and SA-V \cite{ravi2024sam} denote the training video splits of corresponding VOS datasets with mask annotations, while S and V refers to static-image and video training data, respectively. \#L.P. represents the number of learnable parameters. Our M2Pv2 and M2Pv3 are built upon DINOv2 and DINOv3 with specified LoRA ranks (R). The best result for each backbone is shown in bold.}
\label{tab:zero_tapvid_strided}
\end{table*}

\subsection{Comparison with State-of-the-Art VFMs}
Our experiment results comparing VFMs on TAP-VID-DAVIS, TAP-VID-Kinetics and RGB-Stacking are presented in Table \ref{tab:zero_tapvid_strided}. 
Our M2Pv2 and M2Pv3 models are built upon DINOv2 and DINOv3 backbones with specific LoRA ranks (R = 64 or 32).
By performing M2P learning, M2Pv2-S/14 and M2Pv2-B respectively outperform their baselines DINOv2-S and DINOv2-B by large margins of 11.7\% and 12.8\%. Similarly, M2Pv3-B and M2Pv3-S achieve notable gains of 14.6\% and 12.5\% over their corresponding backbones. These results demonstrate that our proposed M2P significantly  improves dense point matching, even when trained with only 3.4K VOS videos. Consistent improvements across various VFM baselines are further observed on TAP-VID-Kinetics in Table \ref{tab:zero_tapvid_strided}.

Compared with zero-shot VFMs  trained on static images (e.g., CLIP \cite{clip} and SD \cite{stablediffusion}), our M2P-trained models significantly outperform them, showing the effectiveness of our proposed mask-guided temporal learning.
Compared with the video-diffusion-based tracker DiffTrack, our M2Pv3-B/16 achieves better performance on both DAVIS and Kinetics while leveraging a smaller overall input feature size.
 Compared with fully-supervised Chrono that uses synthetic data, M2P achieves superior performance, as our proposed mask-based objectives allow VFMs to learn dense tracking from real videos, avoiding the test-time domain gap. Interestingly, we observe that our performance on RGB-Stacking \cite{doersch2022tap} is inferior to Chrono. This is because RGB-Stacking is a synthetic dataset, which is more close to the synthetic Kubric training dataset. In addition, our M2Pv3-B/16 consistently outperforms Chrono-S/16 on both TAP-VID-DAVIS and TAP-VID-Kinetics datasets by using fewer learnable parameters, which  demonstrates the effectiveness of our M2P.

 \begin{table*}[t]
\centering
\setlength{\tabcolsep}{4.3pt}
\renewcommand{\arraystretch}{1.3}
\begin{tabular}{l|cccccc|cccccc|cccccc}
\toprule
\multirow{2}{*}{Backbone} &
\multicolumn{6}{c|}{TAP-VID-DAVIS-First} &
\multicolumn{6}{c|}{TAP-VID-Kinetics-First} &
\multicolumn{6}{c}{TAP-VID-DAVIS-Strided} \\
\cmidrule(lr){2-7} \cmidrule(lr){8-13} \cmidrule(lr){14-19}
& $\delta^0$ & $\delta^1$ & $\delta^2$ & $\delta^3$ & $\delta^4$ & $\delta_{avg}^{x}$ 
& $\delta^0$ & $\delta^1$ & $\delta^2$ & $\delta^3$ & $\delta^4$ & $\delta_{avg}^{x}$
& $\delta^0$ & $\delta^1$ & $\delta^2$ & $\delta^3$ & $\delta^4$ & $\delta_{avg}^{x}$ \\
\midrule
ResNet-18~\cite{doersch2023tapir,resnet} & 9.0 & 27.3 & 54.9 & 73.7 & 84.1 & 49.8 & 8.6 & 28.8 & 56.5 & 74.2 & 83.3 & 50.3 & 9.7 & 31.1 & 60.6 & 78.3 & 87.0 & 53.3 \\
TSM-ResNet-18~\cite{doersch2022tap,lin2019tsm} & 7.3 & 23.1 & 46.7 & 66.6 & 79.2 & 44.6 & 7.8 & 28.1 & 55.2 & 71.4 & 80.2 & 48.6 & 8.2 & 26.8 & 53.6 & 73.5 & 83.9 & 49.2\\
CoTracker~\cite{karaev2025cotracker} &\textbf{ 27.8} &46.4& 58.6& 64.1& 68.5& 53.1 & \underline{24.1}& 41.9& 55.1& 62.2& 67.0& 50.1 &\textbf{33.7}& 52.5& 63.4& 68.3& 72.0& 58.0 \\
DINOv2 (ViT-S)~\cite{oquab2023dinov2} & 6.0 & 19.9& 47.5& 73.7& 84.5& 46.3 & 4.2& 13.6& 36.3& 65.9& 79.2& 39.8 & 7.3& 22.7& 52.9& 80.0& 89.5& 50.4  \\
DINOv2 (ViT-B)~\cite{oquab2023dinov2} & 8.9& 24.7& 53.8& 77.0& 87.1& 50.3 & 5.1& 16.0& 40.0& 67.9& 80.4& 41.9 & 10.0& 28.6& 59.5& 82.9& 87.4& 54.4 \\
\hline
Chrono (ViT-S) \cite{chrono} & 24.0& \underline{49.2}& \underline{71.2}& \underline{82.8}& \underline{87.9}& \underline{63.0} & \textbf{24.8}& \underline{46.2}& \underline{65.8}& \underline{77.5}& \underline{83.1}& \underline{59.5} & 29.7 & \underline{56.4}& \underline{76.7}& \underline{86.8}& \underline{90.9}& \underline{68.0}\\
\textbf{M2P-Chrono (ViT-S)} & \underline{24.9}& \textbf{51.4}& \textbf{73.4}& \textbf{83.4}& \textbf{88.7}& \textbf{64.4}  & 23.8 & \textbf{46.8} & \textbf{67.7} & \textbf{79.3} & \textbf{85.3} & \textbf{60.6} &\underline{30.5} & \textbf{58.3} & \textbf{78.5} & \textbf{87.6} & \textbf{91.7} & \textbf{69.3}  \\
\bottomrule
\end{tabular}
\caption{\textbf{Backbone comparison on the TAP-Vid datasets}~\cite{doersch2022tap}. For fair comparison, all backbones are evaluated at a final feature map resolution of \textbf{64×64} following \cite{chrono}, \textbf{without applying any iterative refinement}. For each column, the best results are shown in bold.}
\label{tab:davis_kinetics}
\end{table*}

\begin{table}[t]
\vspace{-0.3cm}
\centering
\setlength{\tabcolsep}{4.5pt}
\renewcommand{\arraystretch}{1.3}
\begin{tabular}{lcccccc}
\toprule
\multirow{2}{*}{Backbone} &
\multicolumn{6}{c}{TAP-VID-Kinetics-Strided} \\
\cmidrule(lr){2-7}
& $\delta^0$ & $\delta^1$ & $\delta^2$ & $\delta^3$ & $\delta^4$ & $\delta_{avg}^{x}$  \\
\midrule
ResNet-18~\cite{doersch2023tapir,resnet} & 10.5& 35.3& 65.7& 81.3& 88.6& 56.3 \\
TSM-ResNet-18~\cite{doersch2022tap,lin2019tsm} & 9.15& 33.2& 64.6& 79.2& 86.5& 54.5  \\
CoTracker~\cite{karaev2025cotracker} & \underline{31.0} & 31.0& 50.5& 70.6& 74.9& 58.2  \\
DINOv2 (ViT-S)~\cite{oquab2023dinov2} & 4.9& 15.8& 41.9& 73.8& 86.0& 44.5  \\
DINOv2 (ViT-B)~\cite{oquab2023dinov2} & 5.9& 18.6& 45.9& 75.7& 86.9& 46.6    \\
Chrono (ViT-S) \cite{chrono} & \textbf{32.2}& \underline{55.2}& \underline{73.9}& \underline{84.0}& \underline{88.7}& \underline{66.8}  \\
\cmidrule(lr){1-7}
\textbf{M2P-Chrono (ViT-S)} & 30.8 & \textbf{55.9} & \textbf{75.6} & \textbf{85.7} & \textbf{90.5} & \textbf{67.7} \\
\bottomrule
\end{tabular}
\caption{\textbf{Backbone comparison on the TAP-Vid-Kinetics dataset~\cite{doersch2022tap}} with the strided query mode.}
\label{tab:kinetics}
\end{table}

\subsection{M2P for Offline Fine-tuning TAP}
\begin{figure}
\vspace{-0.5cm}
\begin{center}
   \includegraphics[width=1.0\linewidth]{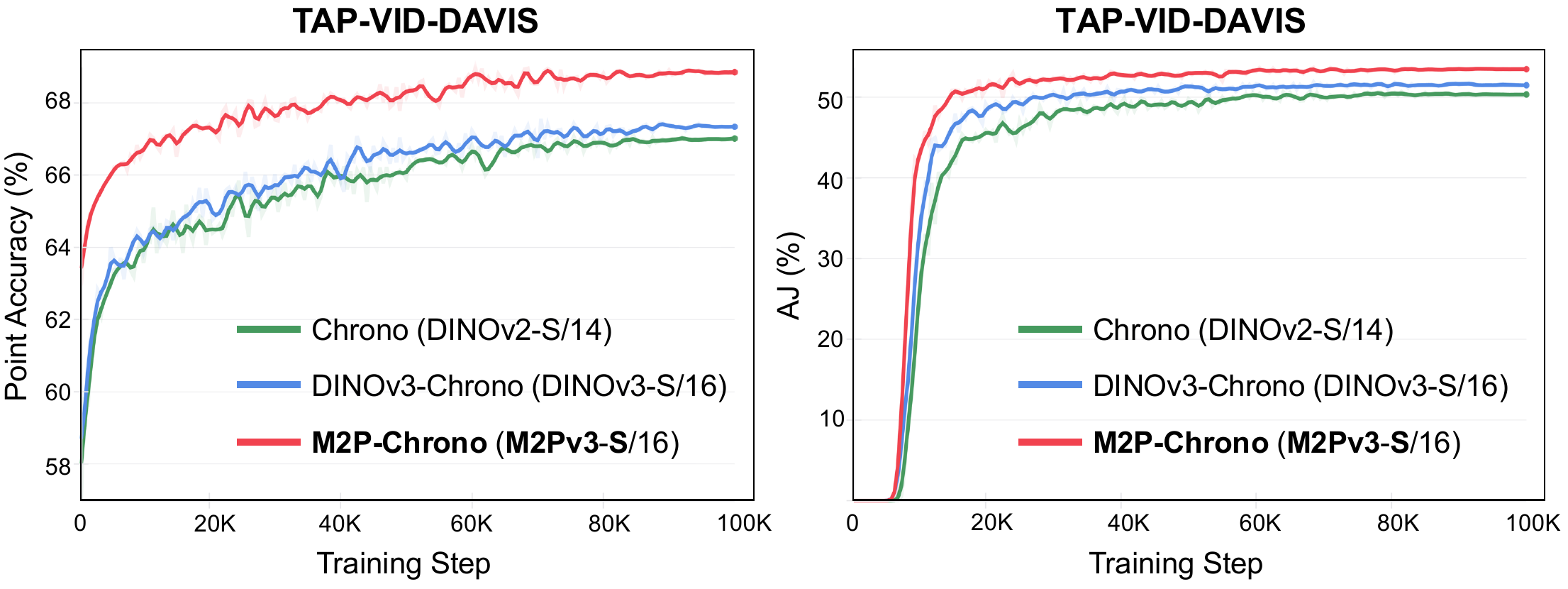} 
\end{center}
\vspace{-0.3cm}
 \caption{\textbf{Training on Kubric \cite{doersch2023tapir} and validation on TAP-VID-DAVIS} with the strided query mode. Our M2P-Chrono significantly outperforms both the original Chrono \cite{chrono} and the Chrono variant with DINOv3-S/16 during the entire training, demonstrating that M2P provides a stronger temporal matching prior, which leads to more stable and efficient fine-tuning on synthetic data.}
\label{training_curves}
\end{figure}

We use our \textbf{M2Pv3-S/16} as the pre-trained model to replace the ViT backbone in Chrono \cite{chrono} for fine-tuning on the synthetic Kubric \cite{doersch2023tapir} dataset. Following Chrono, we learn backbone adapters and adopt the same fine-tuning protocol for a fair comparison. As shown in Tables \ref{tab:davis_kinetics} and \ref{tab:kinetics}, our M2P-Chrono achieves the highest average point accuracy across both TAP-Vid DAVIS and Kinetics datasets under various evaluation modes. This demonstrates that a pre-trained model with stronger temporal priors facilitates temporal correspondence learning from synthetic data, leading to improved tracking performance. Moreover, as shown in Fig.~\ref{training_curves}, we additionally train a Chrono model with the DINOv3-S/16 backbone for fair comparison, following the same training settings as in~\cite{chrono}. Both the original Chrono and the DINOv3-Chrono perform inferior to our M2P-Chrono, demonstrating the superiority of M2P. 
Interestingly, we observe that M2P-Chrono is slightly behind Chrono in the metric $\delta^0$ on Kinetics, which is mainly because M2P-Chrono is built upon DINOv3 with a larger patch size of 16, whereas Chrono uses a patch size of 14, causing minor degradation in extremely fine-grained localization. Finally, M2P-Chrono also outperforms TAP backbones like ResNet18 in \cite{doersch2023tapir} and the transformer in \cite{karaev2025cotracker}.

\subsection{M2P for Test-Time Optimization TAP}
\begin{figure}[t]
\vspace{-0.6cm}
\begin{center}
   \includegraphics[width=0.8\linewidth]{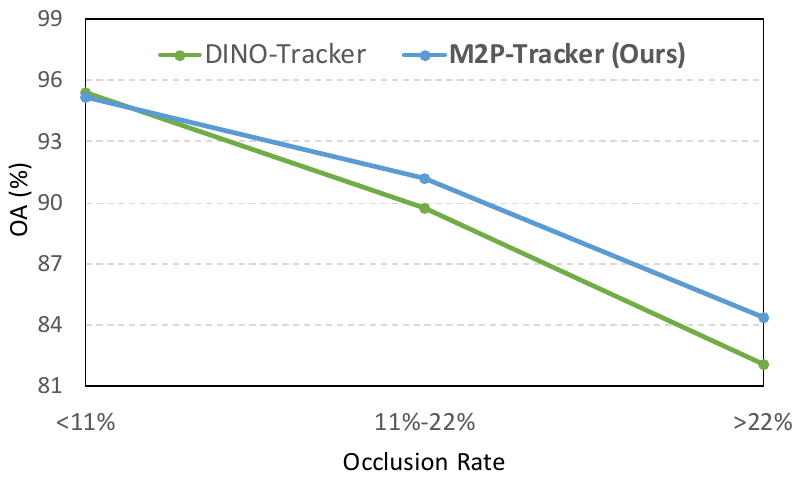} 
\end{center}
\vspace{-0.3cm}
 \caption{Occlusion accuracy (OA) on TAP-Vid-DAVIS-480 \cite{tumanyan2025dino,davis17} under varying occlusion levels.
We partition the TAP-Vid-DAVIS-480 test set into three subsets based on each video’s average occlusion rate, computed from the ground-truth visibility masks.}
\label{m2p_occlusion}
 \vspace{-0.2cm}
\end{figure}

\vspace{-0.05cm}
We replace the backbone in DINO-tracker \cite{tumanyan2025dino} with our \textbf{M2Pv3-S/16} backbone, and adopt LoRA (rank=8) training applied to the last 6 self-attention layers of the backbone for optimization. As shown in Table \ref{tap_davis_480}, with a stronger temporal prior, our M2P-Tracker optimizes \textbf{15$\times$} fewer parameters and achieves a \textbf{22\%} faster training speed compared to the baseline DINO-Tracker, while outperforming it in OA, AJ, and $\delta^{seg}$ metrics. In the comparison on BADJA, M2P-Tracker achieves the leading performance, outperforming supervised offline fine-tuning trackers such as TAPNet and Co-Tracker, which demonstrates the potential of our M2P models for parameter-efficient test-time TAP. In addition, Fig. \ref{m2p_occlusion} further shows that as the occlusion level increases (e.g., $\!>$22\%), our M2P-Tracker shows smaller performance degradation, highlighting its robustness for long-term point tracking under severe occlusions.

\section{Qualitative Visualization}
\label{qualitative_vis_sec}
\begin{figure*}
\begin{center}
   \includegraphics[width=0.95\linewidth]{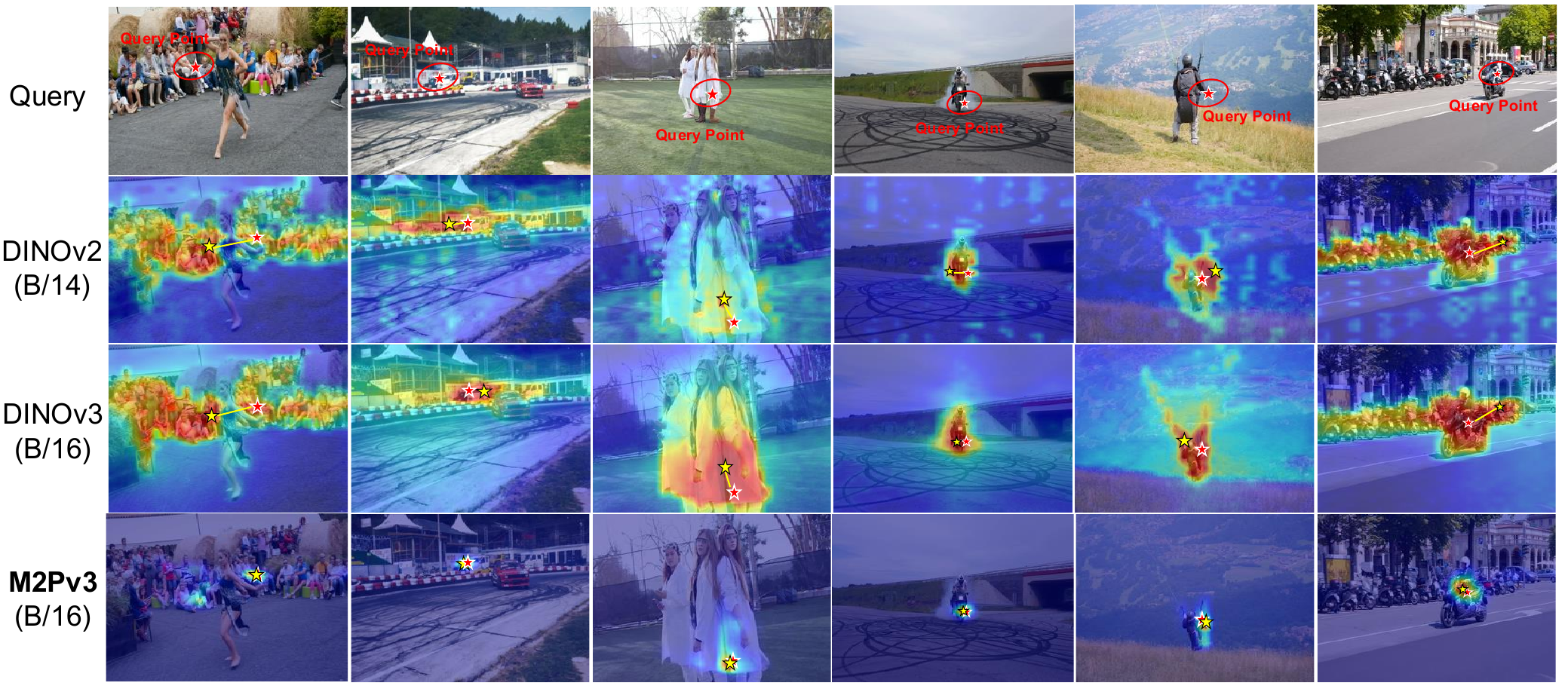} 
\end{center}
\vspace{-0.2cm}
 \caption{Qualitative visualization of \textbf{query-to-frame correlation maps} obtained from DINOv2-B/14 \cite{oquab2023dinov2}, DINOv3-B/16 \cite{dinov3} and our proposed M2Pv3-B/16, where red stars denote query/GT points, and yellow stars denote predicted points. Our M2Pv3 generates sharper and more discriminative correlation maps, while DINOv3 and DINOv2 show noisy responses over distractor regions, which often result in  drifting tracks. Zoom in for clearer visualization.}
\label{qualitative_single}
\end{figure*}

\noindent\textbf{Qualitative results of M2P Backbones.} 
In Fig.~\ref{qualitative_single}, we present the query-to-frame correlation maps produced by DINOv2-B/14~\cite{oquab2023dinov2}, DINOv3-B/16~\cite{dinov3}, and our proposed M2Pv3-B/16. Interestingly, the original DINOv2 and DINOv3 models exhibit strong semantic awareness, often generating high responses over semantically similar but irrelevant regions. Such distractor activations cause the predicted points to drift toward these regions, severely degrading their dense point tracking performance. In contrast, our M2P model effectively suppresses distractor responses and produces sharper, more discriminative correlation maps, leading to substantially more accurate dense point localization.

\begin{figure*}
\begin{center}
   \includegraphics[width=0.95\linewidth]{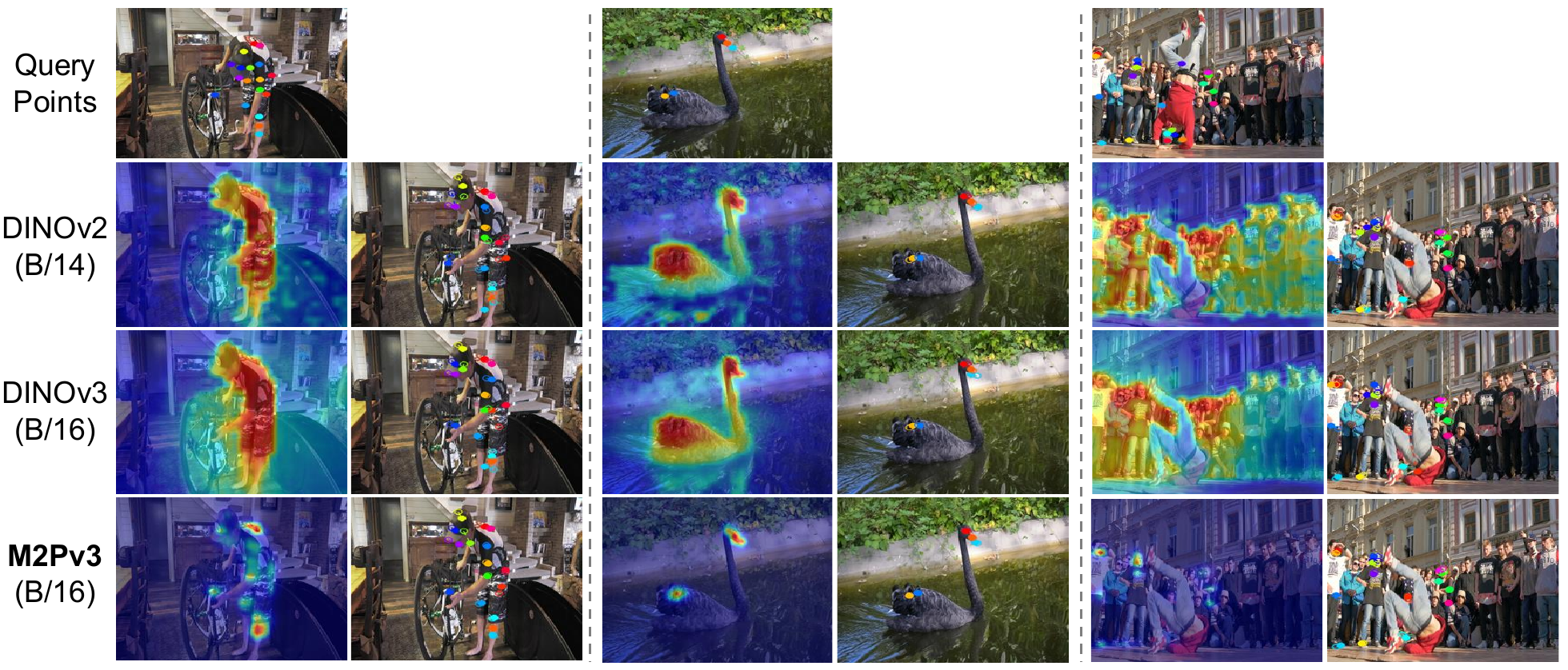} 
\end{center}
\vspace{-0.2cm}
 \caption{Qualitative visualization of the \textbf{average query-to-frame correlation maps} and \textbf{point tracking results}  produced by DINOv2-B/14, DINOv3-B/16, and our M2Pv3-B/16. The average correlation maps are computed by averaging multiple query-to-frame correlation maps. Ground-truth points are shown as open circles, and their corresponding predicted points (i.e., filled circles) are connected with dashed lines to indicate the prediction errors. Please zoom in for clearer details.}
\label{qualitative_all}
\end{figure*}

Fig.~\ref{qualitative_all} shows the overall point tracking results. For completeness, we also visualize the average correlation maps, obtained by averaging multiple query-to-frame correlation maps. As verified in Fig.~\ref{qualitative_single}, DINOv2 and DINOv3 exhibit dense activations over same-class objects, leading to ambiguous and often inaccurate point predictions. In contrast, with the proposed M2P learning, M2Pv3 produces sharper correlation responses while effectively suppressing distractor regions, demonstrating the superiority of our M2P models in dense point tracking.

\begin{table}[t]
\vspace{-0.3cm}
  \newcommand{\tabincell}[2]
    \centering
  \footnotesize
\renewcommand{\arraystretch}{1.3}
\resizebox{\linewidth}{!}{%
    \begin{tabular}{@{\hspace{0.01cm}}c@{\hspace{0.15cm}}c@{\hspace{0.15cm}}c@{\hspace{0.15cm}}c|@{\hspace{0.15cm}}c@{\hspace{0.15cm}}c@{\hspace{0.15cm}}c@{\hspace{0.15cm}}c}
    \Xhline{\arrayrulewidth}
    \multirow{2}{*}{Method}& \multirow{2}{*}{Mode} &  
  \multirow{2}{*}{\#L.P.}& \multirow{2}{*}{Hours} &\multicolumn{3}{c}{DAVIS-480}&\multicolumn{1}{c}{BADJA}\\
 & & & &  \multicolumn{1}{c}{$\delta^{x}_{avg}$} &\multicolumn{1}{c}{OA} &\multicolumn{1}{c}{AJ} & \multicolumn{1}{c}{$\delta^{seg}$} \\
     \hline
     RAFT \cite{raft} & - & - & -  & 66.7 & - & - & 45.0 \cr
     DINOv2 \cite{oquab2023dinov2} & - & -  & -  & 66.7 & - & - & 62.8     \cr  
    \hline
     TAPNet \cite{doersch2022tap} & S & -  & -  & 66.4 & 79.0 & 46.0 & 45.4 \cr
     PIPs++ \cite{zheng2023pointodyssey} & S & -  & -  & 73.6 & - & - & 59.0 \cr
     TAPIR \cite{doersch2023tapir} & S & -  & - & 77.3 & 89.5 & 65.7 & 68.7 \cr
     Co-Tracker \cite{karaev2025cotracker} & S & -  & -.& 79.4 & 89.5 & 65.6 & 64.0 \cr
     \hline 
     Omnimotion \cite{wang2023tracking} & TT & \underline{2.6}M   & 12.5 & 74.1 & 84.5 & 58.4 & 45.2 \cr
     DINO-Tracker \cite{tumanyan2025dino} & TT & 7.6M & \underline{0.9}  & {\textbf{80.4}}  & {\underline{88.1}} & {\underline{{64.6}}} & \underline{72.4} \cr
     \textbf{M2P-Tracker} & TT & \textbf{0.5}M & \textbf{0.7}   & {\underline{79.1}} & {\textbf{89.8}}& {\textbf{65.4}} & \textbf{72.9} \cr
\Xhline{\arrayrulewidth}
   \end{tabular}}
   \centering
    \caption{{Online optimization-based tracker comparison on TAP-Vid-DAVIS-480 \cite{doersch2022tap} and BADJA \cite{biggs2018creatures}}.  Our \textbf{test-time online optimized M2P-Tracker} outperforms DINO-Tracker by using {$15\times$} fewer learnable parameters. `\#L.P.' denotes learnable parameters. Modes `S' and `TT' indicate supervised and test-time optimization, respectively. `Hours' denotes the optimization time per video for each test-time model optimized on an RTX4090 GPU.}
  \label{tap_davis_480}
\end{table}



\begin{figure*}
\begin{center}
   \includegraphics[width=0.95\linewidth]{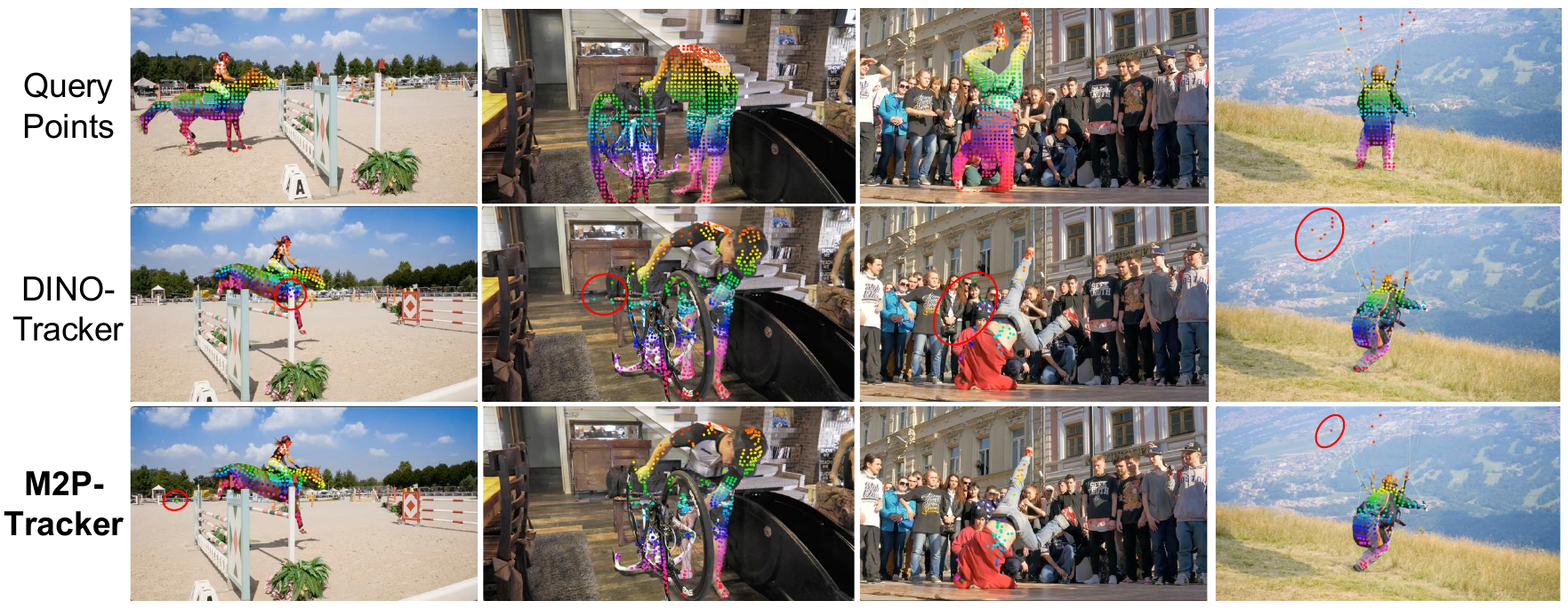} 
\end{center}
\vspace{-0.3cm}
 \caption{Qualitative visualization of the \textbf{dense point tracking results}  obtained by DINO-Tracker \cite{tumanyan2025dino} and our M2P-tracker on TAP-VID-DAVIS 480. We use red circles to highlight the regions with inaccurate tracked points.}
\label{qualitative_m2p_tracker}
\end{figure*}

\noindent\textbf{Qualitative Results of Test-Time Optimization TAP.} The qualitative point tracking results of our test-time optimized M2P-Tracker are illustrated in Fig.~\ref{qualitative_m2p_tracker}.
Our M2P-Tracker is substantially less sensitive to distractors, whereas DINO-Tracker often drifts toward distractor regions or even background points. A possible explanation is that the point features extracted by our M2P backbone can more effectively suppress distractor responses (see Fig.~\ref{qualitative_single}), leading to more stable tracking with fewer drift-caused failures.

\noindent\textbf{Demo Videos for M2P-Tracker.} In the demo videos, we show more qualitative point tracking results of our proposed M2P-Tracker. The results demonstrate that our test-time optimized M2P-Tracker can effectively handle various challenges. Note that our M2P-Tracker only optimizes $15\times$ fewer learnable parameters and achieves $22\%$ training speedup over the baseline DINO-Tracker.



\subsection{Ablation Study}
In this section, we conduct ablation studies to provide more detailed analysis of the proposed M2P learning.

\begin{figure}[t]
\vspace{-0.5cm}
\begin{center}
   \includegraphics[width=1.0\linewidth]{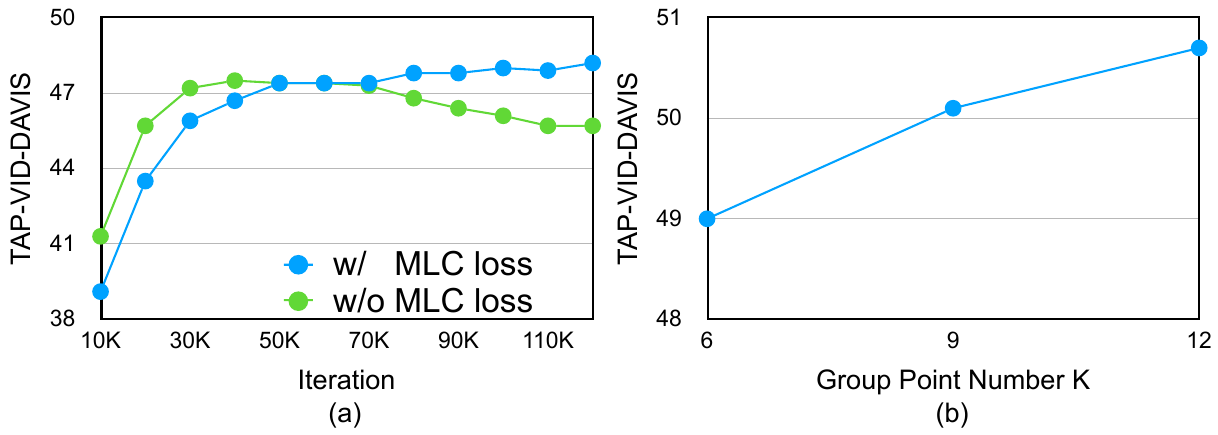} 
\end{center}
\vspace{-0.2cm}
 \caption{(a) Impact of MLC on M2Pv2-B/16-R@8 (QKV); (b) Ablation on the point number $K$ within each group.}
\label{abl_figs}
\end{figure}

\noindent\textbf{Component Analysis.} In Table \ref{abl:comp_analysis}, we evaluate the impact of the three mask-based constrains used in M2P learning. LSC provides explicit point-to-point matching supervision via (\ref{eq:lsc_loss}), which facilitates the VFMs to learn reliable point correspondences, thus improving DINOv2-B/14 by 11.0\%. In addition, the proposed MLC enforces strict foreground point matching, effectively preventing drift toward distractors and enabling more stable  correspondence learning (see Fig.~\ref{abl_figs}(a)). 
MBC further improves the overall performance by using additional near-boundary point supervision.
Overall, the ablation studies demonstrate that M2P can be effectively applied to improve VFMs.


\begin{table}[t]
\vspace{-0.2cm}
  \newcommand{\tabincell}[2]
    \centering
    \setlength{\tabcolsep}{5pt}
\renewcommand{\arraystretch}{1.3}
  \small
    \begin{tabular}{c|@{\hspace{0.5cm}}c@{\hspace{0.6cm}}c@{\hspace{0.6cm}}c@{\hspace{0.5cm}}|c}
    \Xhline{\arrayrulewidth}
  VFM &LSC &MLC & MBC & $\delta^{x}_{avg}$ \cr
       \Xhline{\arrayrulewidth}
   DINOv2-B/14 & \xmark & \xmark & \xmark & 38.0 \cr
   DINOv2-B/14 & \cmark & \xmark & \xmark & 49.0 (+\textbf{11.0\%}) \cr
   DINOv2-B/14 & \cmark & \cmark & \xmark & 50.0 (+\textbf{12.0\%}) \cr
    DINOv2-B/14 & \cmark & \xmark & \cmark & 50.3 (+\textbf{12.3\%}) \cr
   \rowcolor{gray!6}
   DINOv2-B/14 & \cmark & \cmark & \cmark & \textbf{50.8} (+\textbf{12.8\%}) \cr
   \hline
   DINOv3-S/16 & \xmark & \xmark & \xmark & 38.0 \cr
   DINOv3-S/16 & \cmark & \xmark & \xmark & 48.9 (+\textbf{10.9\%})  \cr
   DINOv3-S/16 & \cmark & \cmark & \xmark & 49.6 (+\textbf{11.6\%}) \cr
   \rowcolor{gray!6}
   DINOv3-S/16 & \cmark & \cmark & \cmark & \textbf{50.5} (\textbf{+12.5\%}) \cr
   \Xhline{\arrayrulewidth}  
   \end{tabular} 
   \centering
  \caption{Component analysis of our M2P learning on TAP-VID-DAVIS with the first query mode.}  
  \label{abl:comp_analysis}
\end{table}

\noindent\textbf{Effect of LoRA Training.} 
We adopt LoRA~\cite{hu2022lora} to efficiently fine-tune VFMs and study different ranks and target modules. Here, {QKV} denotes the \emph{Query}, \emph{Key}, and \emph{Value} projections in self-attention, while {Proj.} and {FC} indicate the output projection and feed-forward layers. In Table~\ref{tab:lora_ablation}, applying LoRA to QKV with a larger rank ($r=64$) delivers the strongest point-tracking performance. Extending LoRA to the Projection and Fully-Connected layers yields further improvements, surpassing the supervised TAPNet while using \textbf{20.8\%} fewer learnable parameters (\#L.P).

\begin{table}[t]
\vspace{-0.2cm}
\centering
\small
\setlength{\tabcolsep}{5pt}
\renewcommand{\arraystretch}{1.3}
\resizebox{\linewidth}{!}{
\begin{tabular}{c@{\hspace{0.5cm}}c@{\hspace{0.6cm}}c@{\hspace{0.6cm}}c|@{\hspace{0.5cm}}c}
 \Xhline{\arrayrulewidth}
Variant & Module & Rank & \#L.P. & $\delta^{x}_{avg}$ \\
\Xhline{\arrayrulewidth}
DINOv2-B/14 & - & - & - &  38.0 \\
DINOv2-B/14 & QKV & 8 & 0.3M & 48.3 \\
DINOv2-B/14 & QKV & 16 & 0.6M & 49.0 \\
DINOv2-B/14 & QKV & 32 & 1.2M & 49.4 \\
DINOv2-B/14 & QKV & 64 & 2.4M &\underline{50.5} \\
DINOv2-B/14 & QKV & 128 & 4.7M & 50.1 \\
\rowcolor{gray!6}
DINOv2-B/14 & QKV+Proj.+FC & 64 & 9.5M & \textbf{50.8} \\
\hline
\textcolor{gray}{TAPNet (Supervised)} & \textcolor{gray}{-} & \textcolor{gray}{-} & \textcolor{gray}{12.0M} & \textcolor{gray}{48.6} \\
 \Xhline{\arrayrulewidth}
\end{tabular}}
\caption{
Ablation of LoRA training on TAP-VID-DAVIS with the first query mode.
}
\label{tab:lora_ablation}
\end{table}

\noindent\textbf{Effect of Pre-trained VFMs.} 
In Table~\ref{tab:zero_tapvid_strided}, we apply M2P to different pre-trained VFMs. Stronger VFMs consistently yield stronger M2P models: M2Pv3-B/16 improves over M2Pv2-B/14 by 2.7\%, and M2Pv3-S exceeds M2Pv2-S by 1.7\%. This is mainly because more capable VFMs generate more reliable $\tilde{\bp}$ in (\ref{point_set}), leading to more accurate point transformations.


\noindent\textbf{Pseudo Point Learning.} In our LSC loss, we estimate reliable point correspondences for motion transformation. Here we use these pseudo point-wise correspondences to directly train the VFM. As reported in Table~\ref{abl:more_variants}, this pseudo-point learning variant yields a 7.6\% improvement over the baseline, yet remains inferior to our M2P. This is mainly because M2P leverages more reliable transformed points for learning, leading to more stable and accurate learning.


\noindent\textbf{Effect of Group Point Number $K$.} We evaluate the number of sampled points per group in Fig.~\ref{abl_figs}(b). As observed, increasing the number of sampled points generally improves performance (e.g., $K=12$), which is mainly due to more reliable pseudo point selection for transformation estimation and richer supervision during learning.

\begin{table}[t]
\vspace{-0.3cm}
      \setlength{\tabcolsep}{7pt}
\renewcommand{\arraystretch}{1.3}
  \newcommand{\tabincell}[2]
    \centering
    \begin{tabular}{c@{\hspace{0.5cm}}|c@{\hspace{0.5cm}}}
    \Xhline{\arrayrulewidth}
  Variant & $\delta^{x}_{avg}$ \cr
       \Xhline{\arrayrulewidth}
   DINOv2-B/14 & 38.0 \cr
    w/ Pseudo Point Learning &45.6 \cr
       \rowcolor{gray!6}
    w/ M2P Learning &\textbf{50.8} \cr
    \Xhline{\arrayrulewidth}  
   M2Pv2-B/14 & 50.8 \cr
      \rowcolor{gray!6}
   w/ More Data (D + Y + MOOSE \cite{ding2023mose}) & \textbf{51.5} \cr
       \Xhline{\arrayrulewidth}  
   M2Pv3-S/16 & 50.5 \cr
      \rowcolor{gray!6}
   w/ More Data (D + Y + MOOSE \cite{ding2023mose}) & \textbf{51.2} \cr
   \Xhline{\arrayrulewidth}  
   \end{tabular} 
   \centering
  \caption{Component analysis of our M2P learning on TAP-VID-DAVIS with the first query mode.}  
  \label{abl:more_variants}
\end{table}

\noindent\textbf{More Training Data.} To verify whether our M2P models can benefit from more training data, we further include the MOSSE~\cite{ding2023mose} dataset with 3.6K videos, jointly training the models with the original datasets (D + Y). As shown in Table~\ref{abl:more_variants}, M2P achieves consistent improvements, demonstrating that more training data is beneficial for M2P learning.




\section{Limitation and Future Work}
\label{limitation_fu_work_sec}
In this work, we apply LoRA training to enable efficient M2P learning using a relatively small set of training videos (D + Y). To obtain more discriminative M2P models, it would be better to relax these constraints by increasing the model capacity and leveraging larger-scale VOS training datasets (e.g., SA-V \cite{ravi2024sam}). Moreover, as shown in Fig.~\ref{qualitative_all}, using raw backbone features alone still leads to failures under distractors and severe target deformations. Addressing these challenges may require incorporating online memory mechanisms to enable more robust and adaptive target modeling during tracking.

\section{Conclusion}
In this work, we present Mask-to-Point (M2P) learning, a new framework that leverages object-level mask annotations from video object segmentation datasets to enhance VFMs for dense point tracking through weakly-supervised learning.
M2P introduces mask-guided learning objectives that include a local structure consistency loss and a mask label consistency loss for learning reliable correspondences, together with a boundary supervision loss that  constrains points near object boundaries.
Experiments demonstrate that M2P significantly outperforms existing VFMs on TAP-Vid benchmarks, showing its potential to serve as a strong pre-trained backbone for  TAP.


\section*{Acknowledgments}
This work is partially supported by NSF, ONR, and the Simons Foundation. 
This work is also partially supported by a Strategic Research Grant from
City University of Hong Kong (Project. No. 7005995).

\section{Supplementary}
\input{M2P_supp}

\bibliographystyle{IEEEtran}
\bibliography{main}

\vfill

\end{document}

%% file: M2P_supp.tex
\hyphenation{op-tical net-works semi-conduc-tor IEEE-Xplore}



\title{Supplementary Material\\M2P: Improving Visual Foundation Models with Mask-to-Point Weakly-Supervised Learning for Dense Point Tracking}

\author{Qiangqiang~Wu,
        Tianyu~Yang,
        Bo~Fang,
        Jia~Wan,
        Matias~Di~Martino,
        Guillermo~Sapiro,~\IEEEmembership{Fellow,~IEEE,}
        and~Antoni~B.~Chan,~\IEEEmembership{Senior Member,~IEEE}
\thanks{Q. Wu, B. Fang, and Antoni B. Chan are with the Department of Computer Science, City University of Hong Kong, China (e-mail: qiangqwu2-c@my.cityu.edu.hk; bofang6-c@my.cityu.edu.hk; abchan@cityu.edu.hk). Q. Wu is also with the Department of Electrical and Computer Engineering, Princeton University, Princeton, NJ 08544 USA.}
\thanks{T. Yang is with Meituan, Shenzhen, China. (e-mail: tianyu-yang@outlook.com)}
\thanks{J. Wan is with the School of Computer Science and Technology, Harbin Institute of Technology, Shenzhen 518066, China (e-mail:
jiawan1998@gmail.com).}
\thanks{Matias Di Martino is with the Department of Electrical and Computer Engineering, Duke University, Durham, NC 27708 USA.}
\thanks{Guillermo Sapiro is with the Department of Electrical and Computer Engineering, Princeton University, Princeton, NJ 08544 USA.}
}




\maketitle

In this Supplementary, we detail the estimation steps of transformation parameters via Procrustes Analysis. 
Specifically, as illustrated in (5) of the main paper, given the top-$K_e$ confident correspondence pairs 
$\mathcal{P}_g = \{(\mathbf{p}_i^g,\tilde{\mathbf{p}}_i^g)\}_{i=1}^{K_e}$,
where $\mathbf{p}_i^g$ denotes the query point in the template frame and 
$\tilde{\mathbf{p}}_i^g$ denotes its predicted point in the target frame, 
we can estimate a similarity transformation parameterized by rotation $\mathbf{R}_{g}\in\real^{2\times2}$, translation $\mathbf{t}_{g}\in\real^{2}$, and isotropic scale $s_{g}\in\real$ that aligns the template points to their predicted target locations, 
\begin{equation}
\min_{s_g,\mathbf{R}_{g},\mathbf{t}_{g}}
\sum_{i=1}^{K_c}
\bigl\|\tilde{\mathbf{p}}_i^g - (s_g \mathbf{R}_{g} \mathbf{p}_i^g + \mathbf{t}_{g})\bigr\|^2.
\end{equation}
The above optimization can be solved via Procrustes analysis \cite{schonemann1966generalized}, and we detail the Procrustes analysis steps in the following. Specifically, we first compute the centroids of points in the template and target frames by
\begin{equation}
\boldsymbol{\mu}_{\tilde P} 
= \frac{1}{K_e}\sum_{i=1}^{K_e}\tilde{\mathbf{p}}_i^g,
\qquad
\boldsymbol{\mu}_{P}
= \frac{1}{K_e}\sum_{i=1}^{K_e}\mathbf{p}_i^g, 
\end{equation}
The centered coordinates are computed as 
\begin{equation}
\tilde{\mathbf{p}}_i^{g\,\prime} 
= \tilde{\mathbf{p}}_i^g - \boldsymbol{\mu}_{\tilde P},
\qquad
\mathbf{p}_i^{g\,\prime} 
= \mathbf{p}_i^g - \boldsymbol{\mu}_{P}.
\end{equation}
To estimate the rotation parameter, we let 
\begin{equation}
\tilde P' = 
[\tilde{\mathbf{p}}_1^{g\,\prime},\dots,\tilde{\mathbf{p}}_{K_c}^{g\,\prime}],
\quad
P' = [\mathbf{p}_1^{g\,\prime},\dots,\mathbf{p}_{K_c}^{g\,\prime}].
\end{equation}
The cross-covariance matrix is obtained by 
\begin{equation}
H = \tilde P' (P')^\top,
\end{equation}
with $\mathrm{SVD}(H)=U\Sigma V^\top$, the optimal rotation is
\begin{equation}
\mathbf{R}_g^\star =
V
\begin{bmatrix}
1 & 0\\[2pt]
0 & \operatorname{sign}(\det(VU^\top))
\end{bmatrix}
U^\top,
\end{equation}
where $\det(\cdot)$ denotes the matrix determinant and $\operatorname{sign}(\cdot)$ is the sign function used to avoid reflections. The optimal scale and translation parameters are  estimated as
\begin{equation}
s_g^\star =
\frac{\operatorname{trace}(\Sigma)}
     {\sum_{i=1}^{K_e}\|\mathbf{p}_i^{g\,\prime}\|_2^2}, \quad
\mathbf{t}_g^\star 
= \boldsymbol{\mu}_{\tilde P} 
  - s_g^\star \mathbf{R}_g^\star \boldsymbol{\mu}_{P}.
\end{equation}

The above formulation introduces the differentiable Procrustes 
implementation used in our method for transformation estimation. The closed-form solution is obtained by using 3 pairs of points, i.e.,  $K_e=3$, making it efficient for estimation.


